\documentclass[a4paper]{llncs}

\usepackage{alltt}
\usepackage{amssymb}
\usepackage{caption}
\usepackage{paralist}
\usepackage{nicefrac}
\usepackage{booktabs}
\usepackage{xspace}
\usepackage{amsmath}
\usepackage{marvosym}
\usepackage{wasysym}
\usepackage{verbatim}
\usepackage{float}
\usepackage{hyperref}
\usepackage{multirow}
\usepackage{cite}
\usepackage[capitalize]{cleveref}
\usepackage[per-mode=symbol,detect-all]{siunitx}
\usepackage{graphics}
\usepackage{amssymb}
\usepackage{enumitem}
\usepackage[table]{xcolor}
\usepackage{booktabs}
\usepackage{colortbl}
\usepackage{ifthen}
\usepackage{mathdots}
\usepackage{arydshln}
\usepackage[outline]{contour}
\usepackage{graphicx}
\usepackage{wrapfig}
\usepackage{subfig}

\definecolor{navy}{RGB}{0,0,128}

\newcommand{\relu}{\text{ReLU}\xspace{}}

\newcommand{\tool}{{\textsc{Cnn-Abs}}}

\newcommand{\lm}{{max_{i=0}^{k-1}\{l_i\}}}
\newcommand{\uf}{{max_{i=0}^{k-1}\{u_i\}}}
\newcommand{\um}{{min\{u_i\ |\ u_i \geq l_{max}\}}}
\newcommand{\gammazero}{{\frac{{-1+\sum_{i=0}^{k-1}{\frac{u_i}{u_i-l_i}}}}{\sum_{i=0}^{k-1}{\frac{1}{u_i-l_i}}}}}
\newcommand{\us}{{max\{u_i\}_{i=0\neq f}^{k-1}}}
\newcommand{\net}{\mathcal{N}} 
\newcommand{\neta}{{\mathcal{N}_{\mathcal{A}}}} 

\newcommand{\R}{\mathbb{R}}
\newcommand{\X}{\mathcal{X}}

\newcommand{\sat}{\texttt{SAT}}
\newcommand{\unsat}{\texttt{UNSAT}}

\newcommand{\mat}[1]{\begin{pmatrix}#1\end{pmatrix}}

\usepackage{tikz,ifthen,pgfplots}
\usetikzlibrary{arrows,trees,backgrounds,automata,shapes,decorations,plotmarks,fit,calc,positioning,shadows,chains}
\usetikzlibrary{quotes,arrows.meta}
\tikzstyle{every pin edge}=[<-,shorten <=1pt]
\tikzstyle{neuron}=[circle,fill=black!25,minimum size=17pt,inner sep=0pt]
\tikzstyle{input neuron}=[neuron, fill=green!50]
\tikzstyle{output neuron}=[neuron, fill=red!50]
\tikzstyle{hidden neuron}=[neuron, fill=blue!50]
\tikzstyle{small neuron}        =[hidden neuron, draw, minimum size=15pt]
\tikzstyle{small input neuron}  =[input neuron , draw, minimum size=15pt]
\tikzstyle{small output neuron} =[output neuron, draw, minimum size=15pt]
\tikzstyle{annot} = [text width=4em, text centered]
\tikzstyle{nnedge} = [-{stealth},shorten >=0.1cm, shorten <=0.05cm,line width=0.8pt,black]
\tikzstyle{edge} = [->,line width = 0.3pt, shorten >=0.2cm]
\tikzstyle{edgeWide} = [->,line width = 2pt, , shorten >=0.2cm]
\usetikzlibrary{calc}

\newcommand{\refsubref}[2]{\ref{#1}~\subref{#1-#2}}

\begin{document}
	
	\title{An Abstraction-Refinement Approach to Verifying
		Convolutional Neural Networks}
	
	\author{
			Matan Ostrovsky\inst{1} \and
			Clark Barrett\inst{2} \and
			Guy Katz\inst{1}
		}
	\institute{
			The Hebrew University of Jerusalem, Jerusalem,
                        Israel \\
                        \{matan.ostrovsky, g.katz\}@mail.huji.ac.il
			\and
			Stanford University, Stanford, USA \\
                        barrett@cs.stanford.edu
		}
	
	\maketitle
	
	\begin{abstract}
		Convolutional neural networks have gained vast popularity due to their excellent performance in the fields of computer vision, image processing, and others. Unfortunately, it is now well known that convolutional networks often produce erroneous results --- for example, minor perturbations of the inputs of these networks can result in severe classification errors. Numerous verification approaches have been proposed in recent years to prove the absence of such errors, but these are typically geared for fully connected networks and suffer from exacerbated scalability issues when applied to convolutional networks. To address this gap, we present here the \tool{} framework, which is particularly aimed at the verification of convolutional networks. The core of \tool{} is an abstraction-refinement technique, which simplifies the verification problem through the removal of convolutional connections in a way that soundly creates an over-approximation of the original problem; and which restores these connections if the resulting problem becomes too abstract. \tool{} is designed to use existing verification engines as a backend, and our evaluation demonstrates that it can significantly boost the performance of a state-of-the-art DNN verification engine, reducing runtime by 15.7\% on average.
	\end{abstract}
	
	\section{Introduction}
	
	In machine learning (ML), we use data demonstrating a system's desired behavior to automatically train an artifact that implements the system. ML has become a leading solution for complex algorithmic problems in recent years, obtaining astounding results in many fields. Perhaps the most popular and successful among ML approaches are those for training \emph{deep neural networks} (\emph{DNN}s) --- artifacts that have demonstrated a remarkable ability to solve extremely complex tasks~\cite{CiMeSc12, SiZi14, SzLiJiSeReAnErVaRa15, KrSuHi12, LaXuLiZh15, GoPoMiXuWaOzCoBe14, HaZhReSu16}.
	
	Despite their excellent performance, DNNs are notoriously opaque to human engineers: attempting to ``connect the dots'' and infer the reasoning learned by the DNN is a herculean task. This opacity is particularly troubling, as various errors have been demonstrated in real-world, state-of-the-art DNNs. Perhaps the most famous among these is the \emph{adversarial input} phenomenon, where slight input perturbations cause the DNN to perform severe errors~\cite{SzZaSuBrErGoFe13}. These errors, and others, are a hindrance to the adoption of DNN-based methods in critical systems~\cite{AmOlStChScMa16} (e.g., autonomous vehicles, banking, financial infrastructure, and others), where it is vital to be confident that the system behaves correctly, even in corner cases.
	
	To address this difficulty and facilitate the adoption of DNNs
        in critical systems, different methods for explaining,
        interpreting, and reasoning about DNNs have been proposed. In
        recent years, the verification community has taken an interest
        in \emph{DNN verification}: developing automated tools for
        determining whether a network satisfies a prescribed property
        or providing a counterexample if it does not. Relevant
        properties include, for example, a network's robustness
        against adversarial inputs~\cite{BaIoLaVyNoCr16}, or the
        absence of bias against various protected groups in its
        decision making~\cite{HaPrSr16}. Unfortunately, the DNN formal
        verification problem is NP-complete even for simple neural
        networks and specifications~\cite{KaBaDiJuKo17, KaBaDiJuKo21},
        and becomes exponentially harder as the network size
        increases. Still, great efforts are being put into devising
        verification schemes that can solve average instances of the
        problem quickly, and which support the verification of
        additional kinds of DNNs and properties~\cite{Eh17, TjXiTe17,
          SiGePuVe19, KaBaDiJuKo17, KaHuIbJuLaLiShThWuZe19, WuZh21,
          YaYaTrHoJoPr21, KoLo18, XuLiZhDu21, TrBaXiJo20,
          BoWeChLiDa19, HuKwWaWu17, KuKaGoJuBaKo18, GeMiDrTsChVe18,
          WaPeWhYaJa18, WeZhChSoHsDaBoDh18, DuJhSaTi18, ZhWeChHsDa18,
          LyKoKoWoLiDa20, LoMa17, AmWuBaKa21, ChNuRu17, ElGoKa20,
          AmOlStChScMa16, KaBaDiJuKo21, JaBaKa20, AmScKa21, LaKa21,
          HuKrRuShSuThWuYi20, PaWuGrCaPaBaCl21, ZhShGuGuLeNa20,
          AsHaKrMo20, StWuZeJuKaBaKo21, KaBaDiJuKo17Fvav}.
	
	Here, we contribute to this ongoing effort and present a new framework called \tool{}, which uses an \emph{abstraction-refinement} based approach for verifying \emph{convolutional neural networks} (\emph{CNNs}). CNNs are a particular type of DNNs, which use \emph{convolutions}: constructs that allow for a very compact representation of the DNN, and consequently enable engineers to overcome memory-related bottlenecks. CNNs have been shown to perform well in image processing and computer vision tasks~\cite{SiZi14, SzLiJiSeReAnErVaRa15, KrSuHi12, HaZhReSu16} and are in widespread use. Existing verification tools can verify CNNs, but typically only by reducing them to the general, \emph{fully connected} case, thus failing to leverage the built-in compactness of CNNs. Because the size of the DNN slows down its verification, such transformations are costly. In contrast, our proposed framework aims to utilize the special properties of a CNN in expediting its verification.
	
	At a high level, given a verification query over a CNN, \tool{} first creates an \emph{abstract} network, with significantly fewer neurons, such that proving the property on this smaller network would directly imply that the property also holds for the original network. Notably, the abstract network that we construct is fully connected --- i.e., not convolutional --- and can thus be verified using existing technology. Further, because the verification complexity depends on the number of neurons and edges in the DNN, verifying this smaller network is faster than transforming the CNN into an equivalent, fully connected network and verifying it, as is usually done. Due to the abstraction procedure, verifying the smaller network might produce a spurious counterexample, in which case our framework refines the network and repeats the process.
	
	Apart from the convolution construct, CNNs also make extensive use of \emph{max-pooling} layers. An additional contribution that we make here is proposing a new way for analyzing these layers, which allows us to prune the search space of the resulting verification problem and thus accelerate the verification procedure even further. Specifically, we propose a way to put linear approximate bounds on the max function, which improves the state-of-the-art.
	
	For evaluation purposes, we created a proof-of-concept implementation of our framework. This implementation is comprised of a set of Python modules and is designed to allow for seamless integration with existing DNN verifiers as black-box backends. For the experiments reported in this paper, we used the Marabou DNN verifier~\cite{KaHuIbJuLaLiShThWuZe19} as a backend. We conducted experiments comparing the performance of \tool{} with Marabou as a backend to those of vanilla Marabou and found that the abstraction-refinement approach for verifying CNNs indeed offers significant performance improvements --- specifically, reducing the runtime by 15.7\% on average, with a median runtime reduction of 24.6\%. We also saw an increase of 13\% in the number of realistically solvable queries compared to vanilla Marabou. These results showcase the high potential of \tool{}.
	
	The rest of the paper is organized as follows. In Section~\ref{section:Preliminaries} we provide background on DNNs and their verification. In Section~\ref{section:Abstraction} we schematically present our suggested abstraction-refinement approach, and then discuss our treatment of max-pooling layers in Section~\ref{section:boundPropegation}. We discuss the implementation details of \tool{} in Section~\ref{section:Implementation}, and present its evaluation in Section~\ref{section:Evaluation}. We refer to related work in Section~\ref{section:Related Work}, and conclude in Section~\ref{section:Conclusion}.
	
	\section{Preliminaries}
	\label{section:Preliminaries}
	
	\subsection{Deep Neural Networks}
	
	\begin{figure}[h]
		\begin{center}
			\subfloat[][]{\scalebox{0.8}{
					\begin{tikzpicture}[scale=1]
						\tikzset{>=latex}
						
						\def \dx{1};
						\def \dy{1.2};
						
						\def \xla{0};
						\def \xlb{2};
						\def \xlc{4};
						\def \xld{6};
						\def \xle{8};
						\def \xlf{10};
						
						\def \yla{0};
						\def \ylb{0.6};
						\def \ylc{0.6};
						\def \yld{1.8};
						\def \yle{1.8};
						\def \ylf{0.6};

						\def \nnla{4};
						\def \nnlb{3};
						\def \nnlc{3};
						\def \nnld{1};
						\def \nnle{1};
						\def \nnlf{3};
						
						\def \titleDist{2cm}
						\def \activationDist{1cm}
						
						\def \wdth{0.2pt};
						
						\node at (\xla ,4) [anchor=south, above=\titleDist] {\large Input};
						\node at (\xlb ,4) [anchor=south, above=\titleDist, align=left] {\large Conv.};
						\node at (\xlc ,4) [anchor=south, above=\titleDist, align=left] {\large \relu{}};
						\node at (\xld ,4) [anchor=south, above=\titleDist, text width=1.5cm] {\large Max-Pooling};
						\node at (\xle ,4) [anchor=south, above=\titleDist] {\large WS(1)};
						\node at (\xlf ,4) [anchor=south, above=\titleDist, text width=1.5cm] {\large WS(2) \& Output};
						
						\foreach \ybconst in {0,...,\nnlb}{
							\foreach \yk in {0,1}{
								\node at (\xla, \yla + \ybconst*\dy + \yk*\dy) [circle, fill=black, minimum size=2pt,
								inner sep=0pt, outer sep=0pt] {};
								\ifthenelse{\yk = 0}{
									\draw [edge] (\xla, \yla + \ybconst*\dy + \yk*\dy) -- (\xlb, \ylb + \ybconst*\dy) node[near end, below] {\normalsize $-1.3$};
								}{
									\draw [edge] (\xla, \yla + \ybconst*\dy + \yk*\dy) -- (\xlb, \ylb + \ybconst*\dy) node[near end, above] {\normalsize $1$};
								}
							}
						}
						\foreach \yb in {0,...,\nnlb}{
							\node at (\xlb, \ylb + \yb*\dy) [circle, fill=black, minimum size=2pt,
							inner sep=0pt, outer sep=0pt] {};
							\draw [edge](\xlb, \ylb + \yb*\dy) -- ( \xlc , \ylb + \yb * \dy);
						}
						\foreach \ydconst in {0,1}{
							\foreach \yc in {0,1}{
								\node at (\xlc, \ylc + 2*\ydconst*\dy + \yc*\dy) [circle, fill=black, minimum size=2pt, inner sep=0pt, outer sep=0pt] {};
								\draw [edge] (\xlc, \ylc + 2*\ydconst*\dy + \yc*\dy) -- (\xld, \yld + 2*\ydconst*\dy - 0.5);
							}
						}
						\foreach \yd/\ye/\val in {0/0/1, 0/1/-1, 1/0/3, 1/1/2}{
							\node at (\xld, \yld + 2*\yd*\dy - 0.5) [circle, fill=black, minimum size=2pt, inner sep=0pt, outer sep=0pt] {};
							\ifthenelse{{\ye=0}}{
								\draw [edge]( \xld , \yld + 2*\yd*\dy - 0.5) -- ( \xle , \yle + 2*\ye*\dy - 0.5) node[near end, below, xshift=-3pt] {\normalsize  ${\val}$};
							}
							{
								\draw [edge]( \xld , \yld + 2*\yd*\dy - 0.5) -- ( \xle , \yle + 2*\ye*\dy - 0.5) node[near end, above, xshift=-3pt] {\normalsize ${\val}$};
							}
						}
						\foreach \ye/\yf/\val in {0/0/1, 0/1/-1, 0/2/3, 1/1/2, 1/3/1}{
							\node at ( \xle , \yle + 2*\ye*\dy - 0.5) [circle, fill=black, minimum size=2pt, inner sep=0pt, outer sep=0pt] {};
							\ifthenelse{{\ye=0}}{
								\draw [edge]( \xle , \yle + 2*\ye*\dy - 0.5) -- ( \xlf , \ylf + \yf * \dy) node[midway, below, xshift=0pt] {\normalsize ${\val}$};
							}
							{
								\draw [edge]( \xle , \yle + 2*\ye*\dy - 0.5) -- ( \xlf , \ylf + \yf * \dy) node[midway, above, xshift=0pt] {\normalsize ${\val}$};
							}
						}
						
						\foreach \y/\index in {0/4,1/3,2/2,3/1,4/0} {
							\node at (\xla, \yla + \y*\dy) [small input neuron] {\normalsize $x_{\index}$};
						}
						\foreach \y/\b/\index in {0/0.2/3, 1/0.2/2, 2/0.2/1, 3/0.2/0} {
							\node at (\xlb, \ylb + \y*\dy) [small neuron] {\normalsize $c_{\index}$};
							\node at (\xlb, \ylb + \y*\dy + 0.45) {\normalsize ${\b}$};
						}
						\foreach \y/\index in {0/3 , 1/2, 2/1, 3/0} {
							\node at (\xlc, \ylc + \y*\dy) [small neuron] {\normalsize $r_{\index}$};
							\node at (\xlc, \ylc + \y*\dy + 0.45) {\normalsize $\relu{}$};
						}
						\foreach \y/\index/\below in {0/1/-0.45,1/0/0.45} {
							\node at (\xld, \yld + 2*\y*\dy - 0.5) [small neuron] {\normalsize $m_{\index}$};
							\node at (\xld, \yld + 2*\y*\dy - 0.5 + \below) {\normalsize $Max$};
						}
						\foreach \y/\b/\index in {0/-3/1, 1/5/0} {
							\node at (\xle, \yle + 2*\y*\dy - 0.5) [small neuron] {\normalsize $f_{\index}$};
							\node at (\xle, \yle + 2*\y*\dy - 0.5 + 0.45) {\normalsize ${\b}$};
						}
						\foreach \y/\index/\b in {0/3/0, 1/2/-12, 2/1/2, 3/0/10} {
							\node at (\xlf, \ylf + \y*\dy) [small output neuron] {\normalsize $y_{\index}$};
							\node at (\xlf, \ylf + \y*\dy + 0.45) {\normalsize ${\b}$};
						}
					\end{tikzpicture}
					\hspace{-1.0cm}	}	
				\setlength{\tabcolsep}{5pt}
				\label{fig:toyCNN-a}
			}\vspace{0pt}
			\subfloat[][]{$
				\overset{Input}{\mat{1\\0\\1\\0\\0}} \overset{Conv}{\implies} \mat{1.2\\-1.3\\1.2\\0.2} 
				\overset{\relu{}}{\implies} \mat{1.2\\0\\1.2\\0.2} 
				\overset{\begin{subarray}{l}
						Max-\\
						Pooling
				\end{subarray}}{\implies} \mat{1.2\\1.2} 
				\overset{WS1}{\implies} \mat{6.2\\1.8} 
				\overset{WS2}{\implies} 
				\overset{Output}{\mat{16.2\\7.4\\-1.4\\1.8}}
				$
				\label{fig:toyCNN-b}
			}
		\end{center}
		\caption{(a) A toy CNN demonstrating different layer types. Layer types appear above the layers, activation functions and biases appear above the neurons, and the weights appear above their edges. (b) An evaluation example for the DNN in (a).}
		\label{fig:toyCNN}
	\end{figure}
	
	A feed-forward deep neural network $\net$ is an acyclic weighted directed graph. It has $n$ inputs and $m$ outputs and can be regarded as a mapping from $\R^n$ to $\R^m$. The nodes of $\net$, also called neurons, are organized into layers: the first layer is the \emph{input layer}, the final one is the \emph{output layer}, and the remaining ones are \emph{hidden layers}. When the DNN is invoked, the input layer is assigned values by the caller. Then, in the following layers, the value of each neuron is computed using values of neurons from preceding layers, eventually producing the assignment of the output layer, which is returned to the caller.
	
	The evaluation of each neuron depends on the type of its layer. We focus here on four popular layer types: \emph{weighted sum}, \emph{convolution}, \emph{\relu{}} and \emph{max-pooling} layers, explained next. The example in Fig.~\refsubref{fig:toyCNN}{a} illustrates these different kinds of layers, and an example of an evaluation of the network appears in Fig.~\refsubref{fig:toyCNN}{b}.
	
	In a \emph{weighted sum} (\emph{WS}) layer, each neuron is computed as a weighted sum of values of neurons from the preceding layer. Let vector $V$ represent the assignment of a weighted sum layer of size $t$, and let $U$ represent the assignment of its preceding layer of size $l$. $V$ is then computed as follows:
	\[
	\forall\ 0 \leq i \leq {t-1}. \quad V[i] = B[i] + \sum_{j=0}^{l-1} W[i,j] \cdot U[j]
	\] 
	where $B$ is the \emph{bias} vector, and $W$ is a weight matrix in which $W[i,j]$ is the weight of the edge between neurons $U[j]$ and $V[i]$. A zero weight $W[i,j]=0$ indicates that the edge in question does not exist. $B$ and $W$ are determined when the DNN is trained, before its evaluation~\cite{FoBeCu16}. WS layers are also called \emph{fully connected}, since every neuron can be connected to each neuron in the preceding layer.
	
	A \emph{convolution layer} is similar to a WS layer, but with additional constraints. Whereas before, each neuron would compute a weighted sum of neurons from the previous layer independently of its siblings, now all neurons in the convolution layer share the same set of bias and weights. This set is called a \emph{kernel}. Typically, in a convolution layer, each neuron only considers a small subset of neurons from the previous layer as part of its computation. Thus, weight matrices associated with convolution layers tend to be highly sparse, whereas the matrices in fully connected layers are typically dense.
	
	Let $V$ denote the assignment of a convolution layer with a kernel of size $k\leq l$, a weight vector $W$ and a bias $b$, and let $U$ denote the assignment of the preceding layer of size $l$. Then $V$ is of length $l-k+1$, and its values are as follows:
	\[
	\forall\ 0 \leq i \leq l-k. \quad V[i] = b +
	\sum_{j=0}^{k-1}W[j] \cdot U[i+j]
	\]
	Notice the critical point that $b$ and $W$ are shared among all neurons in the layer.
	
	In a \emph{\relu{} layer} of size $t$, each neuron applies the piecewise-linear \emph{rectified linear unit} (\relu{}) function on a single neuron from the preceding layer:
	\[
	\forall\ 0 \leq i \leq {t-1}. \quad V[i]=\relu{}(U[i])=max(0,U[i])
	\]
	
	Lastly, a neuron in a \emph{max-pooling layer} applies the max function on a set of neurons from the preceding layer. Like with convolution layers, a neuron is typically connected to a small subset of neurons from the preceding layer, and these sets are typically non-overlapping. The output $V$ of a max-pooling layer, in which every neuron is connected to $k$ neurons from the previous layer, is:
	\[
	\forall\ 0 \leq i \leq \frac{l}{k}-1. \quad V[i] = max_{j=0}^{k-1} \{U[ik + j]\}
	\]
	where $U$ is a vector of size $l$ representing the assignment of the preceding layer.
	
	A convolutional neuronal network is a DNN containing some convolution layers, and typically also max-pooling layers~\cite{SiZi14, KrSuHi12, SzLiJiSeReAnErVaRa15}. The network in Fig.~\refsubref{fig:toyCNN}{a} is an example of a CNN.
	
	\subsection{Formal Verification of Neural Networks}
	
	In formal verification of DNNs, we use automated procedures to check whether a network $\net$ satisfies some desirable specification. A verification \emph{property} has the form $ P(x) \land Q(\net(x)), $ and is comprised of a set $P$ of input constraints, and a set $Q$ of output constraints that encode an \emph{undesired} behavior.  Given a verification query $\langle P,N,Q\rangle$, the verifier's goal is to find a \emph{counterexample} (\emph{CEX}) $x_0 \in \R^n$ such that $P(x_0) \land Q(\net(x_0))$ holds, demonstrating that the undesired behavior is possible. When such an $x_0$ is found, the verifier returns \sat{}, and provides $x_0$; otherwise, it returns \unsat{}, indicating that the DNN behaves as desired. Typically, $P$ and $Q$ are restricted to be conjunctions of linear constraints~\cite{KaBaDiJuKo17}. Multiple sound and complete approaches have been proposed for solving the verification problem (e.g.,~\cite{KaBaDiJuKo17, KaHuIbJuLaLiShThWuZe19, Eh17, TjXiTe17, DuJhSaTi18}).
	
	An example of a property for the network in Fig.~\ref{fig:toyCNN} is: 	\begin{equation}
		\left(\bigwedge_{i=0,2} (0.5 \leq x_i \leq 1) \land \bigwedge_{i=1,3,4} (0 \leq x_i \leq 0.5)\right) \land \left( y_1 \leq y_0 \right) 
		\label{equation:toyQuery}
	\end{equation}
	In this case, a sound verifier would return \sat{}, with a possible counterexample being $x_0 = \langle 1,0,1,0,0\rangle$, for which $y_0=16.2$ and $y_1=7.4$. 
	
	\subsection{Bound Propagation}
	
	Many of the activation functions used in popular DNN architectures are \emph{piecewise linear}; i.e., they can be regarded as having two or more distinct linear output phases, each associated with a specific input region. For example, \relu{} alternates between the identity function for positive inputs and the zero function for non-positive ones. As part of the verification process, verification tools often perform \emph{case splitting}~\cite{KaBaDiJuKo21}; i.e., they guess that a satisfying assignment exists when the piecewise-linear constraint is restricted to one of its linear phases, and then backtrack if that guess turns out to yield an \unsat{} result. This case splitting is often the most costly part of the verification process.
	
	To reduce the number of case splits and thus expedite the verification process, verification engines will often try to deduce that a piecewise-linear function is fixed to one of its linear phases, thus translating a difficult piecewise-linear constraint into a linear one. This deduction is typically performed by computing lower- and upper-bounds for inputs to activation functions, hoping that the entailed range falls entirely within one of the function's linear segments. There exist many methods for such bound computation~\cite{LoMa17, SiGePuVe19, TjXiTe17, Eh17, KaHuIbJuLaLiShThWuZe19, WuZh21, YaYaTrHoJoPr21, TrBaXiJo20, BoWeChLiDa19, GeMiDrTsChVe18, WaPeWhYaJa18, WeZhChSoHsDaBoDh18, DuJhSaTi18, ZhWeChHsDa18, LyKoKoWoLiDa20, SaYaZhHsZh19, AnHuMaTjVi20, AmWuBaKa21}.
	
	For example, observe the network in Fig~\ref{fig:toyCNN}, and the property given in Eq.~\ref{equation:toyQuery}. This property places bounds on $x_0$ and $x_1$: $x_0 \in [0.5,1]$ and $x_1 \in [0,0.5]$. These bounds can be propagated to the neurons in the ensuing convolution layer: it is straightforward to show, using \emph{interval arithmetic}, that $c_0$ is consequently bound to the range $[0.05,1.2]$. Next, $c_0$'s bounds can be propagated through the \relu{} layer, to derive that $r_0\in[0.05, 1.2]$, and is thus restricted to one of its linear phases --- specifically, this \relu{} is just the identity function, $r_0=c_0$.
	
	\section{Abstraction-Refinement of Convolutional Neural Network}
	\label{section:Abstraction}
	
	\subsection{The General Abstraction-Refinement Framework}
	
	The DNN verification problem is hard and costly to solve, especially for CNNs --- where the mere translation into a fully connected network that the verification engine can process might already incur a significant blowup in the number of constraints that need to be encoded into the solver. In \tool{}, we propose a new heuristic that can accelerate this process, and which focuses on applying abstraction/refinement principles~\cite{ClGrJhLuVe00}.
	
\begin{figure}[h]
		\begin{center}
			\subfloat[][$\net$]{
				\begin{tikzpicture}
					\tikzset{>=latex}
					\node [small input neuron] (x) at (0,0) {$x$};
					\node [left=of x, xshift=1cm] (xrange) {$[-1,1]$};
					\node [small neuron] (h0) at (1,0.5) {$h_0$};
					\node [small neuron] (h1) at (2,0.5) {$h_1$};
					\node [small output neuron] (y) at (3,0) {$y$};
					\draw[->] (x)  -- (h0) node[midway, above=2pt] {$1$};
					\draw[->] (x)  -- (y) node[midway, below=2pt]  {$-1$};
					\draw[->] (h0) -- (h1) node[midway, above=2pt] {$1$};
					\draw[->] (h1) -- (y)  node[midway, above=2pt] {$1$};
				\end{tikzpicture}
				\label{fig:absEx-a}
			}
			\hspace{0.5cm}
			\subfloat[][$\neta$]{
				\begin{tikzpicture}
					\tikzset{>=latex}
					\node [small input neuron] (x) at (0,0) {$x$};
					\node [small input neuron] (h1) at (2,0.5) {$h_1$};
					\node [small output neuron] (y) at (3,0) {$y$};
					\draw[->] (x)  -- (y) node[midway, below=2pt]  {$-1$};
					\draw[->] (h1) -- (y)  node[midway, above=2pt] {$1$};
					\node [left=of x, xshift=1cm] (xrange) {$[-1,1]$};
					\node [left=of h1, xshift=1cm] (xrange) {$[-1,1]$};
				\end{tikzpicture}
				\label{fig:absEx-b}
			}
			
		\end{center}
		\caption{A simple network (a), and an even simpler
			network that over-approximates it, where $h_0$ is
			removed (b).}
		\label{fig:absEx}
	\end{figure}
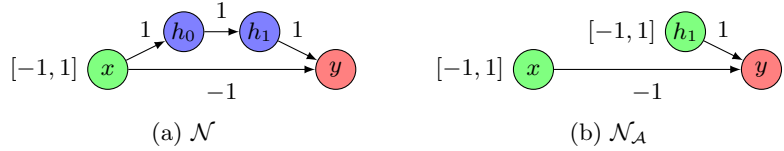

	We start with a motivating example. Observe network $\net$ in Fig.~\ref{fig:absEx}~\subref{fig:absEx-a}, which is comprised strictly of weighted sum layers; and suppose we wish to solve a verification query where the input variable $x$ is restricted to the range $[-1,1]$. It is straightforward to show that $h_0=h_1=x$ and $y = 0$, and that $h_0,h_1 \in [-1,1]$. Now, observe $\neta$ at Fig.~\refsubref{fig:absEx}{b} --- it is identical to $\net$, except that $h_0$ is deleted and $h_1$ is consequently regarded as an input neuron (no incoming edges), with the range $[-1,1]$. Because the range of $h_1$ in $\neta$ is the same as in $\net$, it can be shown that $y$ in $\neta$ can obtain any value that $y$ in $\net$ could obtain. Therefore, if a verification query $\langle P \land (h_1\in[-1,1]),\neta,Q\rangle$ is \unsat{}, then $\langle P,\net,Q\rangle$ is guaranteed to be \unsat{} as well.  In this case, we say that $\neta$ is an \emph{over-approximation}, or an \emph{abstraction}, of $\net$. For example, suppose we set $P=(-1\leq x\leq 1)$ and $Q=(y\geq 5)$. It is straightforward to show that $\langle P\wedge(-1\leq h_1\leq 1), \neta, Q\rangle$ is \unsat{}, because its $y$ can not exceed $2$; and consequently, $\langle P,\net, Q\rangle$ is also \unsat{}.
	
	More formally, we propose the following abstraction framework. Assume that we are given a verification query $\langle P, \net, Q\rangle$, and that our goal is to find an $x$ such that $P(x)\land Q(\net(x))$, or prove that no such $x$ exists. Given a formula $\varphi$, let $\X_{\varphi}=\{x\ |\ \varphi(x)\}$. We construct a new network, $\neta$, with the same output neurons as in $\net$, and an input layer which is a superset of that of $\net$. Additionally, we create a condition $P_B$ which specifies bounds on inputs introduced in the abstraction process, such that $\net(\X_P)\subseteq \neta(\X_{P\land P_B})$. The following lemma establishes that $\neta$ is an over-approximation of $\net$ (with respect to $P$ and $Q$):
	\begin{lemma}
		Let $\langle P,N, Q\rangle$ be a verification query. Let $\neta$ be a DNN with the same output layer as $\net$, and an input layer that is a superset of $\net$'s input layer. Let $P_B$ be a condition over $\net$'s neurons that is logically implied by $P$, such that $\net(\X_P)\subseteq\neta(\X_{P\land P_B})$. If $\langle P\land P_B,\neta,Q\rangle$ is \unsat{}, then $\langle P,\net,Q\rangle$ is also \unsat{}.
		\label{lemma:overApproximation}
	\end{lemma}
	The proof is straightforward, and is omitted. We refer to $\neta$ as an abstraction of $\net$.
	
	Using Lemma~\ref{lemma:overApproximation}, we can use the following scheme for verifying $\langle P, \net, Q\rangle$. First, we solve the query $\langle P\land P_B, \neta, Q\rangle$; if this query is \unsat{}, we can immediately answer that $\langle P, Q\rangle$ is \unsat{} for $\net$. Otherwise, we have counterexample $x \in \X_{P\land P_B}$ such that $\neta(x)$ satisfies $Q$. After removing the assignment of excess input neurons from $x$, we then check whether $\net(x)$ satisfies $Q$, and if so, answer \sat{} and return $x$ as a counterexample for the original query. However, it is possible that that $\net(x)$ violates $Q$ --- in which case we call $x$ a \emph{spurious} counterexample. For example, for $\net$ in Fig.~\ref{fig:absEx}, and property $P=(-1\leq x \leq 1)\land(-1\leq h_1 \leq 1),Q=(y\geq 1)$, we have a counterexample for $\neta$ where $h_1=1,x=1$ and $y=2$, but assigning $x=1$ in $\net$ results in $y=0$.
	
	When the abstraction process yields a spurious counterexample, we can construct a new network, $\neta'$, such that $ \net(\X) \subseteq \neta'(\X_{P\land P'_B}) \subsetneq \neta(\X_{P\land P_B}) $, and $\neta'$'s inputs are a superset of $\net$'s and constrained by $P'_B$. Then, we repeat the process by verifying $\langle P\land P'_B, Q\rangle$ on $\neta'$. The idea is that, while $\neta'$ is still an abstraction of $\net$, it is a \emph{refinement} of $\neta$. Specifically, $\neta'(\X_{P\land P'_B})$ is built as a strict subset of $\neta(X_{P\land P_B})$, hopefully eliminating many spurious counterexamples. Under the reasonable assumption that after sufficiently many refinement steps we will obtain the original network $\net$, this iterative process is bound to terminate --- either by obtaining an \unsat{} result on one of the networks, indicating unsatisfiability of the original query, or by finding a true counterexample (possibly by verifying the original network itself). Of course, an abstraction/refinement scheme is useful only if a conclusion can be reached before refining the network back to the original.
	
	For soundness and completeness, we state the following lemma, whose proof is straightforward and is again omitted:
	\begin{lemma}
		If the abstraction refinement scheme described above uses a sound and complete DNN verifier to dispatch its verification queries, and after finitely many refinement steps we obtain the original network, then the scheme is also sound and complete.
	\end{lemma}
	
	The abstraction/refinement scheme described above is general, and its effectiveness depends greatly on how it is instantiated; specifically, on:
	\begin{inparaenum}[(i)]
		\item how the initial abstraction is generated; and
		\item how each refinement step is performed.
	\end{inparaenum}
	Next, we propose a specific instantiation that is adequate for the CNN setting, as our experiments later demonstrate.
	
	\subsection{\tool{}'s Abstraction-Refinement Scheme}
	\label{section:CNNAbsAbstractionRefinment}
	
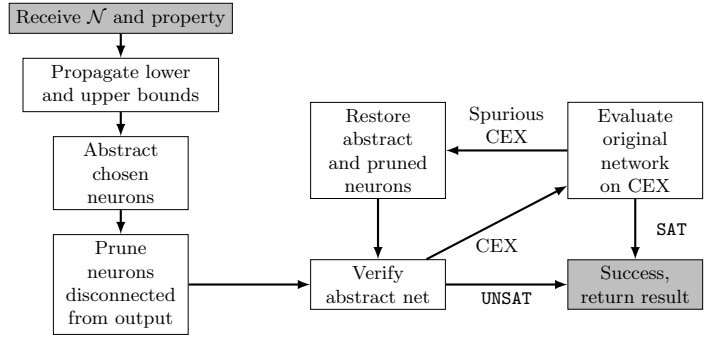
\begin{figure}[htp]
		\centering
		\scalebox{0.8}{
			\begin{tikzpicture}[align=center, node distance=1cm and 2cm]
				\tikzset{>=latex}
				\def \w{2cm};
				\node [draw, rectangle, fill=lightgray] (start) {Receive $\net$ and property};
				\node [draw, rectangle, below=of start, yshift=0.6cm, text width=\w+1cm] (bounds) {Propagate lower and upper bounds};
				\node [draw, rectangle, below=of bounds, yshift=0.6cm, text width=\w] (abstraction) {Abstract chosen neurons};
				\node [draw, rectangle, below=of abstraction, yshift=0.6cm, text width=\w] (prune) {Prune neurons disconnected from output};
				\node [draw, rectangle, right=of prune, text width=\w] (verify) {Verify abstract net};
				\node [draw, rectangle, above=of verify, text
				width=\w] (fail) {Restore abstract and pruned neurons};
				\node [draw, rectangle, right=of fail, text
				width=\w] (inputs) {Evaluate original network
					on CEX};
				\node [draw, rectangle, right=of verify, text width=\w, fill=lightgray] (success) {Success, return result};

				\draw [->, line width=1pt] (start) -- (bounds);
				\draw [->, line width=1pt] (bounds) -- (abstraction);
				\draw [|-,-|,->, line width=1pt] (abstraction) -- (prune);
				\draw [|-,-|,->, line width=1pt] (prune) -- (verify);
				\draw [->, line width=1pt] (verify) -- (success) node[text width=\w, midway,below] {\unsat{}};
				\draw [->, line width=1pt] (verify) -- (inputs) node[text width=\w, midway,below=4pt] {CEX};
				\draw [->, line width=1pt] (inputs) -- (fail) node[text width=\w, midway,above] {Spurious CEX};
				\draw [->, line width=1pt] (inputs) -- (success) node[text width=\w, midway, right=-15pt] {\sat{}};
				\draw [|-,-|,->, line width=1pt] (fail) -- (verify);
			\end{tikzpicture}
		}
		\caption{The suggested abstraction-refinement scheme.}
		\label{fig:abstractionScheme}
	\end{figure}

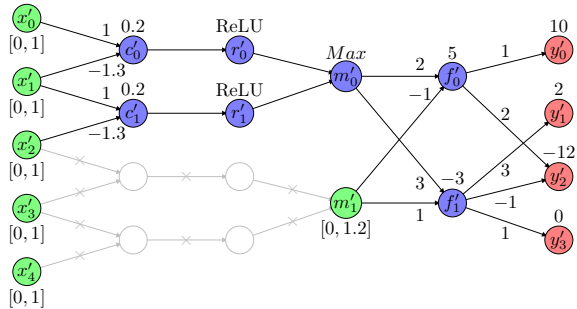
\begin{figure}[h]
			\begin{center}
				\scalebox{0.7}{
					\begin{tikzpicture}[scale=1]
						\tikzset{>=latex}
						
						\def \dx{1};
						\def \dy{1.2};
						
						\def \xla{0};
						\def \xlb{2};
						\def \xlc{4};
						\def \xld{6};
						\def \xle{8};
						\def \xlf{10};
						
						\def \yla{0};
						\def \ylb{0.6};
						\def \ylc{0.6};
						\def \yld{1.8};
						\def \yle{1.8};
						\def \ylf{0.6};

						\def \nnla{4};
						\def \nnlb{3};
						\def \nnlc{3};
						\def \nnld{1};
						\def \nnle{1};
						\def \nnlf{3};
						
						\def \titleDist{2cm}
						\def \activationDist{1cm}
						
						\def \wdth{0.2pt};
						\foreach \ybconst in {0,...,\nnlb}{
							\foreach \yk in {0,1}{
								\ifthenelse{\ybconst < 2}
								{
									\draw [edge, lightgray] (\xla, \yla + \ybconst*\dy + \yk*\dy) -- (\xlb, \ylb + \ybconst*\dy) node[midway] {\normalsize $\times$};
								}
								{
									\node at (\xla, \yla + \ybconst*\dy + \yk*\dy) [circle, fill=black, minimum size=2pt, inner sep=0pt, outer sep=0pt] {};
									\ifthenelse{\yk = 0}{
										\draw [edge] (\xla, \yla + \ybconst*\dy + \yk*\dy) -- (\xlb, \ylb + \ybconst*\dy) node[near end, below] {\normalsize $-1.3$};
									}{
										\draw [edge] (\xla, \yla + \ybconst*\dy + \yk*\dy) -- (\xlb, \ylb + \ybconst*\dy) node[near end, above] {\normalsize $1$};
									}
								}
							}
						}
						\foreach \yb in {0,...,\nnlb}{
							\ifthenelse{\yb < 2}{
								\draw [edge, lightgray](\xlb, \ylb + \yb*\dy) -- ( \xlc , \ylb + \yb * \dy) node[midway] {\normalsize $\times$};
							}
							{
								\node at (\xlb, \ylb + \yb*\dy) [circle, fill=black, minimum size=2pt, inner sep=0pt, outer sep=0pt] {};
								\draw [edge](\xlb, \ylb + \yb*\dy) -- ( \xlc , \ylb + \yb * \dy);
							}
						}
						\foreach \ydconst in {0,1}{
							\foreach \yc in {0,1}{
								\ifthenelse{\ydconst < 1}{
									\draw [edge, lightgray] (\xlc, \ylc + 2*\ydconst*\dy + \yc*\dy) -- (\xld, \yld + 2*\ydconst*\dy - 0.5)node[midway] {\normalsize $\times$};
								}
								{
									\node at (\xlc, \ylc + 2*\ydconst*\dy + \yc*\dy) [circle, fill=black, minimum size=2pt, inner sep=0pt, outer sep=0pt] {};
									\draw [edge] (\xlc, \ylc + 2*\ydconst*\dy + \yc*\dy) -- (\xld, \yld + 2*\ydconst*\dy - 0.5);
								}
							}
						}
						\foreach \yd/\ye/\val in {0/0/1, 0/1/-1, 1/0/3, 1/1/2}{
							\node at (\xld, \yld + 2*\yd*\dy - 0.5) [circle, fill=black, minimum size=2pt,
							inner sep=0pt, outer sep=0pt] {};
							\ifthenelse{{\ye=0}}{
								\draw [edge]( \xld , \yld + 2*\yd*\dy - 0.5) -- ( \xle , \yle + 2*\ye*\dy - 0.5) node[near end, below, xshift=-3pt] {\normalsize  ${\val}$};
							}
							{
								\draw [edge]( \xld , \yld + 2*\yd*\dy - 0.5) -- ( \xle , \yle + 2*\ye*\dy - 0.5) node[near end, above, xshift=-3pt] {\normalsize ${\val}$};
							}
						}
						\foreach \ye/\yf/\val in {0/0/1, 0/1/-1, 0/2/3, 1/1/2, 1/3/1}{
							\node at ( \xle , \yle + 2*\ye*\dy - 0.5) [circle, fill=black, minimum size=2pt, inner sep=0pt, outer sep=0pt] {};
							\ifthenelse{{\ye=0}}{
								\draw [edge]( \xle , \yle + 2*\ye*\dy - 0.5) -- ( \xlf , \ylf + \yf * \dy) node[midway, below, xshift=0pt] {\normalsize ${\val}$};
							}
							{
								\draw [edge]( \xle , \yle + 2*\ye*\dy - 0.5) -- ( \xlf , \ylf + \yf * \dy) node[midway, above, xshift=0pt] {\normalsize ${\val}$};
							}
						}
						\foreach \y/\index in {0/4,1/3,2/2,3/1,4/0} {
								\node at (\xla, \yla + \y*\dy) [small input neuron] {\normalsize $x'_{\index}$};
								\node at (\xla,  \yla + \y*\dy - 0.5) {\normalsize $[0,1]$};
						}
						
						\foreach \y/\b/\index in {0/0.2/3, 1/0.2/2, 2/0.2/1, 3/0.2/0} {
							\ifthenelse{\y < 2}{
								\node at (\xlb, \ylb + \y*\dy) [small neuron, fill=white, draw=lightgray] {};
							}
							{
								\node at (\xlb, \ylb + \y*\dy) [small neuron] {\normalsize $c'_{\index}$};
								\node at (\xlb, \ylb + \y*\dy + 0.45) {\normalsize ${\b}$};
							}
						}
						\foreach \y/\index in {0/3, 1/2, 2/1, 3/0} {
							\ifthenelse{\y < 2}{
								\node at (\xlc, \ylc + \y*\dy) [small neuron, fill=white, draw=lightgray] {};
							}
							{
								\node at (\xlc, \ylc + \y*\dy) [small neuron] {\normalsize $r'_{\index}$};
								\node at (\xlc, \ylc + \y*\dy + 0.45) {\normalsize $\relu{}$};
							}
						}
						\foreach \y in {0,...,\nnld} {
							\ifthenelse{\y < 1}{
								\node at (\xld, \yld + 2*\y*\dy - 0.5) [small input neuron] {\normalsize $m'_1$};
								\node at (\xld, \yld + 2*\y*\dy - 0.5 - 0.5) {\normalsize $[0,1.2]$};
							}
							{	
								\node at (\xld, \yld + 2*\y*\dy - 0.5) [small neuron] {\normalsize $m'_0$};
								\node at (\xld, \yld + 2*\y*\dy - 0.5 + 0.45) {\normalsize $Max$};
							}
						}
						\foreach \y/\b/\index in {0/-3/1, 1/5/0} {
							\node at (\xle, \yle + 2*\y*\dy - 0.5) [small neuron] {\normalsize $f'_{\index}$};
							\node at (\xle, \yle + 2*\y*\dy - 0.5 + 0.45) {\normalsize ${\b}$};
						}
						\foreach \y/\index/\b in {0/3/0, 1/2/-12, 2/1/2, 3/0/10} {
							\node at (\xlf, \ylf + \y*\dy) [small output neuron] {\normalsize $y'_{\index}$};
							\node at (\xlf, \ylf + \y*\dy + 0.45) {\normalsize ${\b}$};
						}
				\end{tikzpicture}}
			\end{center}
			\caption{The network from Fig.~\ref{fig:toyCNN}, abstracted by disconnecting the edges leading to $m_1$ and pruning the neurons no longer connected to the output neurons (in gray). $m'_1$ is now an input neuron, bounded by its previously computed bounds.}
			\label{fig:abstractNet}
	\end{figure}

	Given a CNN $\net$ and a property $P(x) \land Q(\net(x))$ to be verified, \tool{}'s abstraction-refinement scheme is summarized in Fig.~\ref{fig:abstractionScheme}. We start by defining a set of neurons $V$, the \emph{abstract neurons}, for which we compute sound lower- and upper-bounds, as in the example in Fig.~\ref{fig:absEx}. We then create the abstract network $\neta$ by deleting the incoming edges to $V$'s neurons, marking them as input neurons, and adding the bounds computed for them as $P_B$, which is added in conjunction to $P$. We finish the construction by removing --- or \emph{pruning} --- all the hidden neurons in $\neta$ that are no longer connected to the output layer, as illustrated in Fig.~\ref{fig:abstractNet}. By its construction, $\neta$ satisfies Lemma~\ref{lemma:overApproximation}, and is hence an over-approximation of $\net$. Once $\neta$ is constructed, we invoke a backend verifier to dispatch $\langle P\land P_B, \neta, Q\rangle$, and proceed as described earlier. Later, we elaborate on \tool{}'s heuristics for selecting the initial $V$, and for performing a refinement step when a spurious counterexample is found; for now, we assume $V$ is selected arbitrarily, and that a refinement step arbitrarily reinstates previously-removed edges, as well as any previously-pruned neurons that again become connected to the output layer. Equivalently, a refinement step can be regarded as constructing a new abstract network, using a set $V'$ of abstract neurons such that $V' \subsetneq V$; and the set $V \setminus V'$, which is selected heuristically, is the set of neurons that are being restored.
	
	We note that our abstraction technique can also be applied to fully connected networks; however, the number of pruned neurons in the resulting abstract networks is expected to be negligible because of the nature of weighted-sum layers. Consequently, we find that convolutional and pooling topologies are better suited for the technique at the core of
	\tool{}.
	
	\subsection{Heuristics for Abstracting and Refining Neurons}
	\label{section:pruningPolicies}
	
	The effectiveness of abstraction schemes is known to depend significantly on the heuristic used for abstraction and refinement~\cite{ElGoKa20, ClGrJhLuVe00}. As part of \tool{}, we propose several such heuristics.
	
	\subsubsection{Selecting the initial abstraction.}
	To maximize the number of neurons that will be pruned as a result of the abstraction operation, we begin by finding the convolution or max-pooling layer that is deepest in the network, and which does not have a fully connected layer preceding it. The idea, as illustrated in Fig.~\ref{fig:abstractNet}, is to create a cone-shaped set of neurons that can be pruned from the network after the abstraction operation. Once this layer is selected, we perform bound propagation (using any of the many existing techniques, e.g.,~\cite{TjXiTe17,SiGePuVe19,GeMiDrTsChVe18}) and abstract all the neurons in that layer. This constitutes our initial abstraction.
	
	\subsubsection{Refining the abstraction.}
	Whenever \tool{} discovers a spurious counterexample, it performs a refinement step, as outlined in Section~\ref{section:CNNAbsAbstractionRefinment}. Since all abstract neurons are located in the same layer, we maintain a ``score of importance'' for each abstract neuron, and \tool{} picks the neuron with the highest score as the one to refine. The motivation is to identify the neurons most relevant to the query at hand so as to quickly converge to a correct answer. We propose five refinement scoring heuristics, described next. Some of these heuristics make use of the DNN's \emph{test-set}~\cite{FoBeCu16}, which, for our purposes, is regarded as a list of $N$ input points and their corresponding correct labels: $\{(s_i,l_i)\}_{i=0}^{N-1}$, where $s_i$ is an input point and $l_i$ is its label (the set of all $L$ possible labels is denoted $\{0,..., L-1\}$). 
	\begin{enumerate}
        \item \textbf{Centered.} This heuristic assumes that the arrangement of the neurons within the layer matters; this happens, for example, in image recognition, where the edge of the image is often less important than its center~\cite{Le98}. Here, we rank neurons according to their distance from the layer's center, assigning higher scores to neurons closer to the center. Observe a layer $U$ that is regarded as a $D$-dimensional array of dimensions $d_1\times d_2 \times\ldots\times d_D$ (as is often done in convolutional networks), and a neuron $v$ in $U$. Suppose that $v$'s coordinates within $U$'s multi-dimensional array are $(j_v^1,...,j_v^D)$. In this case, $v$'s score is computed by negating its distance from the center point of $U$:
          \[
            score(v) = - \sqrt{\sum_{i=1}^{D} (j_v^i - \lfloor\frac{d_1}{2}\rfloor)^2}
          \]

        \item \textbf{All Samples.} Here we rank neurons by their mean assigned values, averaged over the DNN's test-set: for every neuron $v$, we set 
		\[
		score(v)=\frac{\sum_i v(s_i)}{N},
		\]
		where $v(s)$ is the value assigned to $v$ when the DNN is evaluated on input $s$.  The motivation is that the most important neurons are those that are often assigned large values on inputs from the test-set, which is drawn according to the data distribution that the DNN is expected to encounter after deployment. 
		
		\item \textbf{Sample Rank.} This heuristic assumes that property $P$ is focused on a specific input $x_0$ --- as is the case, for example, with adversarial robustness properties~\cite{KaBaDiJuKo17}, which check that the network assigns a consistent label to a small ball centered around $x_0$. Consequently, the heuristic ranks neurons by their assignment:
		\[
		score(v) = v(x_0)
		\]
		
		\item \textbf{Single Class.} Here we again assume that $P$ is focused on a specific input $x_0$, which is assigned an output label $l$. We then rank the layer's neurons by their mean assigned values over all test-set samples also classified as $l$:
		\[
		score(v)=\frac{\sum_{l_i=l} v(s_i)}{|\{l_i\ |\ l_i=l\}|}
		\]
		The motivation is that when considering $x_0$, the test-set samples belonging to the same class as $x_0$ are the more relevant ones.
		
		\item \textbf{Majority Class Vote.} For a neuron $v$, we first calculate its Single Class score for every class $c^v_j = \frac{\sum_{y_i=l_j} v(x_i)}{|\{y_i\ |\ y_i=l_j\}|}$, and then set
		\[
		score(v) = \|c^v\|_2
		\]
		The idea is that the $l^2$-norm generally elevates vectors with distinct large values over vectors with uniformly distributed values. Interpreting $c^v_j$ as the relevance of a neuron for a single class, this policy identifies neurons that are relevant for multiple classes.
	\end{enumerate}
	
	\section{Propagating Bounds through Max-Pooling Layers}
	\label{section:boundPropegation}
	
	As part of \tool{}'s abstraction scheme, we assume that we can compute sound lower- and upper-bounds for the DNN's neurons. Indeed, it is important to discover bounds that are as tight as possible, in order to rule out as many spurious counterexamples down the line. We propose a new bound derivation method for max-pooling layers, which improves over the state-of-the-art, and consequently boosts the scalability of \tool{}.
	
	Many modern bound tightening methods are based on \emph{linear relaxation}~\cite{Eh17, TjXiTe17, KaHuIbJuLaLiShThWuZe19,SiGePuVe19}, where the DNN verification query is encoded using strictly linear constraints.  The constraints in the input and output property are already linear by definition, and can be encoded directly; as can the weighted sum and convolution layers.  Non-linear activation functions, such as \relu{} or max-pooling, are the main challenge, and these are typically encoded by bounding their output values between approximate linear bounds.  Once the linear relaxation is encoded, it can be solved by a linear programming (LP) solver or can be used to approximate the set of possible outputs of the DNN. Because of the approximate nature of the linear query, the solving procedure in this case is often incomplete.
	
	Linear relaxations can also be used to compute lower and upper bounds for the hidden neurons of the DNN.  This can be achieved, for example, by repeatedly invoking an LP solver, each time asking it to maximize or minimize the value of each neuron in the DNN~\cite{TjXiTe17, Eh17}. The optimal values obtained for each neuron then constitute (sound) upper and lower bounds for these neurons, and many solvers use them to determine when a piecewise-linear constraint has become fixed to a linear phase and can be translated into a linear constraint~\cite{GeMiDrTsChVe18, KaBaDiJuKo21}.  In case the LP solver determines the query is infeasible, the verification query is \unsat{}.
	
	The tightness of the bounds produced by such linear relaxation approaches depends primarily on how tightly they approximate the non-linear activation functions. The \relu{} functions, which are highly popular, have received a great deal of attention~\cite{SiGePuVe19,LyKoKoWoLiDa20}. In contrast, the max function, which is a key component in the max-pooling layers common in leading CNN architectures~\cite{KrSuHi12, SzLiJiSeReAnErVaRa15, SiZi14}, has received only limited attention, and there is currently no agreed-upon ``standard'' relaxation for encoding it. We propose here a novel encoding, which improves over the current state of the art~\cite{Eh17, SiGePuVe19, BoWeChLiDa19}, and yields tighter bounds.
	
	For a max constraint $b = max_{i=0}^{k-1} \{a_i\}$, where each $a_i$ has a lower bound $l_i$ and an upper bound $u_i$, we first find the largest upper bound $u_f=\uf$, the second-largest upper bound $ u_s = \us$, the maximal lower bound $l_{max}=\lm$, and the minimal upper bound that is greater than the maximal lower bound $u_{min} = \um$.
	
	If $u_s < l_{max}$, the max constraint is trivial: it has a single input variable $a_i$ that is clearly greater than all the others, because its lower bound is greater than all other upper bounds. In that case, we accurately transform the max constraint into a linear constraint, $b=a_i$. Otherwise, we use the following approximations:
	\begin{multline}
		\bigwedge_{0\leq j \leq k-1}\left( b \geq a_j \right) \land \bigwedge_{\lambda \in \{l_{max},u_{min}\}}\left( b \leq \lambda + \sum_{i=0}^{k-1} \frac{\relu{}(u_i-\lambda)}{u_i-l_i}(a_i-l_i) \right) \land \\
		\left( b \leq u_f \frac{u_s - l_f}{u_f - l_f} + a_f
		\frac{u_f - u_s}{u_f-l_f} \right)
		\label{equation:lpMaxNew}
	\end{multline}
	
\begin{figure}[htp]
		\begin{minipage}[t]{0.25\textwidth}
			\begin{center}
				\vspace{0pt}
				\scalebox{0.8}{\begin{tikzpicture}[shorten >=1pt,->,draw=black!50, node distance=\layersep,font=\footnotesize]
						\def\layersep{1.1cm}
						\def \baseY{0}
						
						\node[small input neuron] (v00) at (0, \baseY + 4.5) {$x_0$};
						\node[small input neuron] (v01) at (0, \baseY + 3) {$x_1$};
						\node[small input neuron] (v02) at (0, \baseY + 1.5) {$x_2$};
						\node[small input neuron] (v03) at (0, \baseY + 0) {$x_3$};
						
						\node[small neuron] (v10) at (1 * \layersep, \baseY + 3 + 0.75) {$c_0$};
						\node[small neuron] (v11) at (1 * \layersep, \baseY + 1.5 + 0.75) {$c_1$};
						\node[small neuron] (v12) at (1 * \layersep, \baseY + 0 + 0.75) {$c_2$};
						
						
						\node[small neuron] (v20) at (2 * \layersep, \baseY + 3) {$m_0$};
						\node[small neuron] (v21) at (2 * \layersep, \baseY + 1.5) {$m_1$};
						\node[annot] at (2 * \layersep, \baseY + 3.5) {Max};
						\node[annot] at (2 * \layersep, \baseY + 2) {Max};
						
						\node[small output neuron] (y) at (3 * \layersep, \baseY + 2.25) {$y$};
						
						\draw[nnedge] (v00) -- node[above] {$1$} (v10);
						\draw[nnedge] (v01) -- node[above] {$-1$} (v10);
						\draw[nnedge] (v01) -- node[above] {$1$} (v11);
						\draw[nnedge] (v02) -- node[above] {$-1$} (v11);
						\draw[nnedge] (v02) -- node[above] {$1$} (v12);
						\draw[nnedge] (v03) -- node[above] {$-1$} (v12);

						\draw[nnedge] (v10) -- node[above] {} (v20);
						\draw[nnedge] (v11) -- node[below] {} (v20);
						\draw[nnedge] (v11) -- node[above] {} (v21);
						\draw[nnedge] (v12) -- node[below] {} (v21);
						
						\draw[nnedge] (v20) -- node[above] {$1$} (y);
						\draw[nnedge] (v21) -- node[below] {$1$} (y);
				\end{tikzpicture}}
			\end{center}
		\end{minipage} \textcolor{lightgray}{\vline} \begin{minipage}[t]{0.33\textwidth} 
			\vspace{0pt}
			\textbf{LP Query:}\\
			\textbf{Maximize $y$\ s.t.}
			
			$\left.\begin{array}{l}
				-1\leq x_0,x_1 \leq 1 \\
				-2\leq x_2,x_3 \leq 2 \\
			\end{array}\right\} \begin{array}{l} \text{Input} \\ \text{const-} \\ \text{raints}\end{array}$\\
			$\left.\begin{array}{l}
				c_0=x_0 - x_1 \\
				c_1=x_1 - x_2 \\
				c_2=x_2 - x_3 \\	
			\end{array}\right\} \begin{array}{l} \text{2nd layer}\end{array}$\\
		\end{minipage} \hspace{2pt} \begin{minipage}[t]{0.33\textwidth}
			\vspace{0pt}
			$\left.\begin{array}{l}
				m_0 >= c_0,c_1 \\
				m_0 - c_0 - \frac{5}{6}c_1 <= 2.5  \\
				m_0 - \frac{1}{6}c_1 <= 2.5  \\
				m_1 >= c_1,c_2 \\
				m_1 - c_1 - \frac{1}{8}c_2 <= 3.5  \\
				m_1 - \frac{1}{8}c_2 <= 3.5  \\
			\end{array}\right\} \begin{array}{l} \text{Max} \\ \text{func-} \\ \text{tions}\end{array}$\\
			$\left.\begin{array}{l}
				y=m_0 + m_1 \\
			\end{array}\right\} \begin{array}{l} \text{4th layer}\end{array}$\\
			\\
			\textbf{Result: maximal y is 6.5}
		\end{minipage}
		\caption{A simple CNN (left) and its encoding as an LP query (right).}
		\label{fig:LPExample}
	\end{figure}
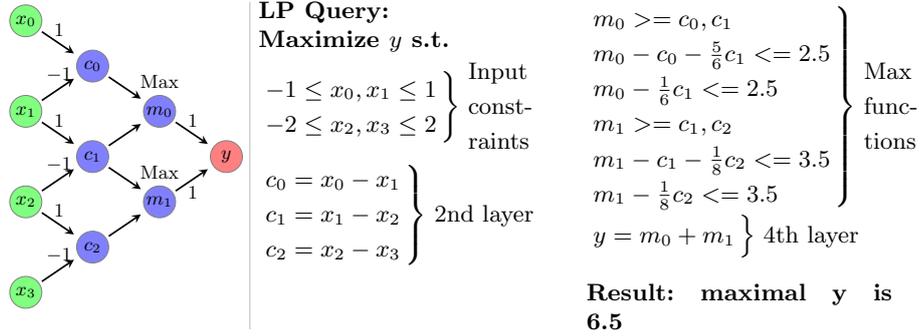

	Additional details, a proof of correctness, and a comparison to the state-of-the-art are provided in Appendix~\ref{section:appendixLP}. We demonstrate our encoding in Fig.~\ref{fig:LPExample}. The simple CNN depicted therein is translated to an LP query. An LP solver is then invoked to maximize $y$ --- in order to obtain an upper bound for it. In this case, a na\"ive, interval propagation of the input bounds would yield the bounds $c_0 \in [-2,2], c_1 \in [-3,3], c_2 \in [-4,4]$, then $m_0 \in [-2,3], m_1 \in [-3,4]$, and finally $y \in [-5,7]$. The LP relaxation yields a tighter upper bound of $y \leq 6.5$. Current state-of-the-art techniques (described thoroughly in Appendix~\ref{section:appendixLP}, Eq.~\ref{equation:stateoftheartLP}), yield a looser upper bound of $y \leq 7$.
	
	\section{Implementation}
	\label{section:Implementation}
	
	We created a proof-of-concept, Python implementation of \tool{}, to be released with the final version of this paper. This tool implements our abstraction/refinement framework, and can be configured to use a black-box DNN verifier and a black-box bound propagation engine as backends. Thus, the tool will benefit from future improvements in bound propagation techniques and verification technology for non-convolutional networks
	
	For input, \tool{} currently accepts CNNs stored in Tensorflow~\cite{TF15} format --- specifically, as Keras~\cite{CH15} sequential objects.  The tool's main module, \emph{CnnAbs.py}, implements the abstraction and refinement policies described in Section~\ref{section:pruningPolicies}. It offers specialized support for adversarial robustness properties~\cite{BaIoLaVyNoCr16}, which comprised our evaluation (see Section~\ref{section:Evaluation}). The central features it includes are:
	
	\textbf{The \emph{CnnAbs} class}, which implements \tool{}'s main functionality, manages solving, logging, and other configurations. It includes the following methods:
	\begin{itemize}		
		\item \emph{solveAdversarial(model, abstractionPolicy, sampleIndex, distance)}: solves an adversarial robustness query on the Keras.Sequential DNN \emph{model}, allowing input perturbations in an $\|\|_{\infty}$-ball of radius \emph{distance} around input sample whose index is \emph{sampleIndex} in the data-set. The abstraction policy used is \emph{abstractionPolicy}. The method returns the \sat{} or \unsat{} results, along with a counterexample for the \sat{} case.
		
		\item \emph{solve(model, modelTF, abstractionPolicy, property)}: solves \emph{model}, which encodes both network and property, using the abstraction policy \emph{abstractionPolicy}. For technical reasons, this method also receives a property object \emph{property} and a Keras sequential model \emph{modelTF}. The method returns the result and possibly a counterexample, and supports general properties beyond adversarial robustness.
		
		\item \emph{propegateBounds(model)}: propagates lower and upper bounds for all neurons in the network and property encoded in \emph{model}.
	\end{itemize}
	
	\textbf{The \emph{ModelUtils} class}, which implements utilities for the Tensorflow interface. Its \emph{tf2Model(model)} method translates a Tensorflow model to a model object. We follow here a common convention~\cite{KaHuIbJuLaLiShThWuZe19} where the network, propagated bounds, and linear property are all stored as a single model.
	
	\textbf{Policy classes:} abstraction policies are implemented as classes inheriting from the \emph{PolicyBase} class. Every child class is required to implement the \emph{rankAbsLayer(model, prop, absLayerPredictions)} function. Its arguments are \emph{model}, a property described in \emph{prop}, and the assigned values of the abstracted layer for each point in the test-set. It returns the variable indices of the layer's neurons, sorted by their score: the first element is the least important and will thus be refined last. This modular design allows adding additional heuristics easily.
	
	\textbf{The \emph{AdversarialProperty} class}, which defines an adversarial robustness property.
	
	\textbf{The \emph{DataSet} class}, which contains the relevant data of the data-set in use.
	
	For bound computation, we used an existing module from the
        Marabou project~\cite{KaHuIbJuLaLiShThWuZe19, WuZeKaBa22, WuOzZeIrJuGoFoKaPaBa20} and extended it with our new bounds for max-pooling layers. That module is implemented in the \emph{LPFormulator.cpp} file.
	
	\section{Evaluation}
	\label{section:Evaluation}
	
	\subsubsection{Network architecture.}
	For evaluation purposes, we trained three convolutional networks on the MNIST digit recognition data-set~\cite{Le98}. The input vectors to these networks represent $28\times 28$ grayscale images, with input pixels restricted to the range $[0,1]$. The first network, network A, has two \emph{convolution blocks} (a convolution layer followed by a \relu{} layer and a max-pooling layer), another block consisting of a weighted-sum layer and a \relu{} layer, and a final weighted-sum layer. When transformed into an equivalent, fully connected model, it has a total of 2719 neurons and achieves a test-set accuracy of 93.7\%. The second network, B, has the same layer sequence as A, but its convolution kernels are larger; consequently, it has 4564 neurons and achieves an accuracy of 96.2\%. Network C is similar but has three convolution blocks instead of two; it has 4636 neurons and achieves an accuracy of 86.6\%. Additional details appear in Appendix~\ref{section:appendixNetworkStructure}.
	
	\subsubsection{Adversarial robustness.}
	We focus here on adversarial robustness properties~\cite{SzZaSuBrErGoFe13, BaIoLaVyNoCr16, CaDaKoKoKiSt21, PaWuGrCaPaBaCl21}, which have become the de-facto standard benchmarks for DNN verification~\cite{SiGePuVe19, Eh17, TjXiTe17, SiGePuVe19, DuJhSaTi18, AmWuBaKa21}. An adversarial robustness query consists of input $x^0$ to some classification DNN $\net$ with $m$ outputs; its goal is to prove that small perturbations to $x_0$ within an $\varepsilon$ ball (for some $\varepsilon>0$) do not result in a change of classification. For simplicity, we consider \emph{targeted} adversarial robustness, where the goal is to prove that some perturbation cannot result in the input being classified as some target label $l$. We select $l$ as the label that received the second-highest score when the DNN is evaluated on $x_0$.
	
	More formally, let $j^0 = argmax_j \{\net(x^0)_j\}$ and $j_s^0 = argmax_{j\neq j^0} \{\net(x^0)_j\}$ denote the maximal and the second-best predicted classes for $x^0$. The adversarial robustness property $\varphi$ for an input $x^0$ and maximal perturbation $\varepsilon$ is:
	\[
	\varphi(x^0, \varepsilon, x) = \left(\|x^0 - x\|_{\infty} \leq \varepsilon\right) \land \left( \net(x)_{j^0} \leq \net(x)_{j^0_s} \right)
	\]
	Other norms, beyond $l^\infty$, could also be used.
	
	\subsubsection{Tool setup.} We used our implementation of \tool{}, with the Marabou DNN verification engine~\cite{KaHuIbJuLaLiShThWuZe19} as a backend. Marabou is a modern DNN verifier that incorporates SMT-based solving~\cite{KaBaDiJuKo17}, abstract interpretation techniques~\cite{SiGePuVe19}, and also uses the Gurobi LP solver~\cite{gurobi} to dispatch LP queries as part of bound computation. All experiments were run with a 1-hour timeout, and individual verification queries on abstract networks were limited to $800$ seconds.
	
	\subsection{Experiments}
	\label{section:expSetup}
	
	\subsubsection{Comparing ranking policies.}
	\label{section:rankPolicies}
\begin{figure}
		\begin{center}
			\subfloat[][]{\includegraphics[width=0.5\textwidth]{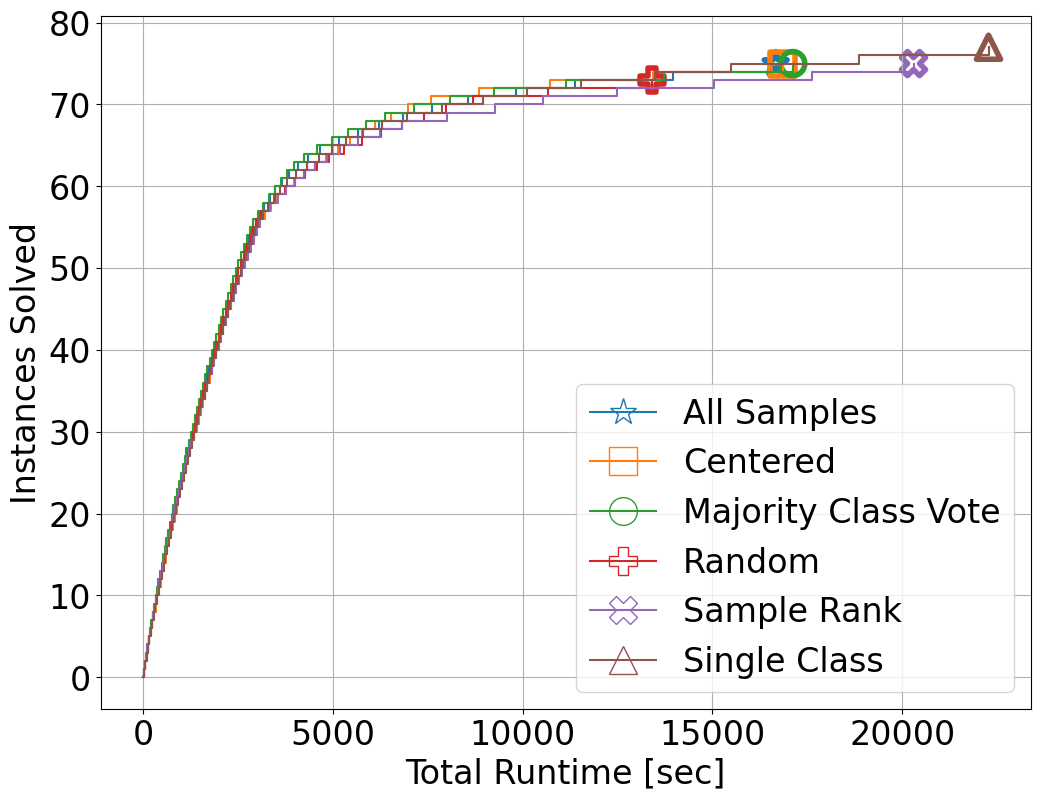}\label{fig:comparePolicies-a}}
			\subfloat[][]{\includegraphics[width=0.5\textwidth]{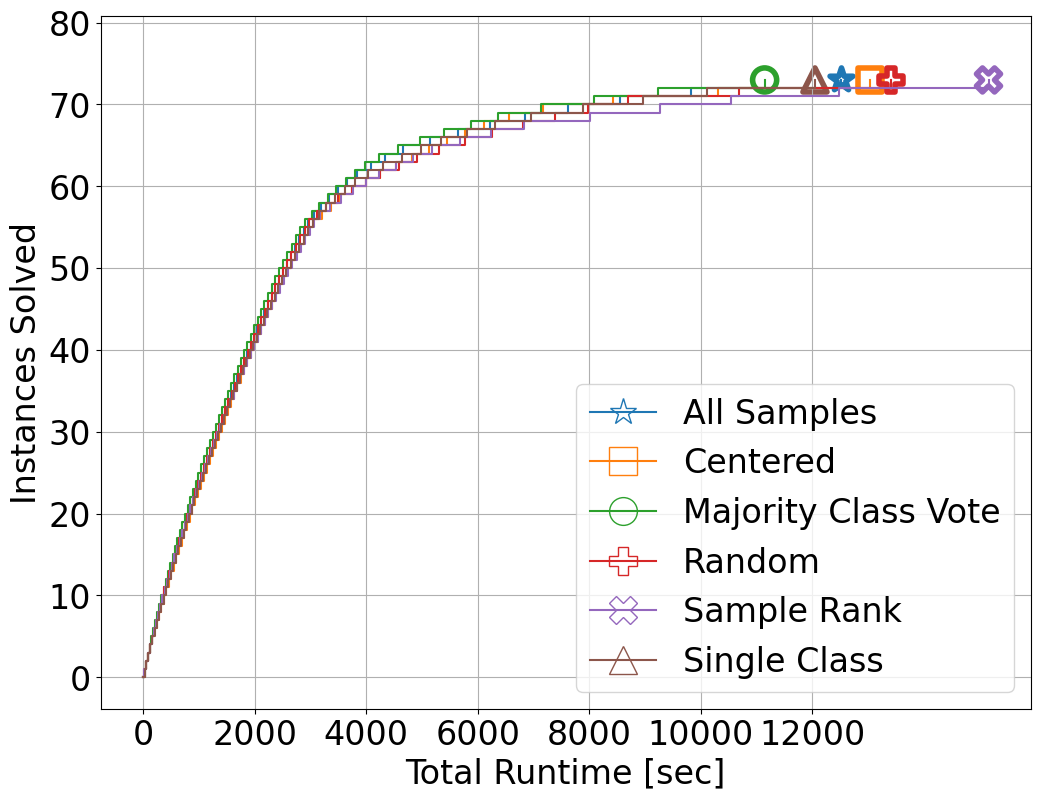}\label{fig:comparePolicies-b}}
		\end{center}
		\caption{(a) Policy comparison on network A, with 			$\varepsilon= 0.03$. (b) The same results,  			restricted to the 73 benchmarks successfully solved by all policies.}
		\label{fig:comparePolicies}
	\end{figure}

	For our first experiment, we set out to evaluate \tool{}'s performance using each of the different ranking heuristics described in Section~\ref{section:pruningPolicies}. To this end, we prepared adversarial robustness queries for the first 100 images from the MNIST test-set, each with an arbitrary $\varepsilon$ value. We then ran \tool{} in each configuration, and also vanilla Marabou, on all 100 benchmarks on network A. For this experiment, the abstraction was performed on the first \relu{} layer of the network, which was large enough to allow the different policies to exhibit their different behaviors. As a control group, we also included a \emph{Random} policy, which ranks neurons in the layer randomly. The results appear in Fig.~\ref{fig:comparePolicies}{a}, which depicts the accumulated number of solved benchmarks as a function of the time passed. Fig.~\ref{fig:comparePolicies}{b} depicts the same results, but it only includes those benchmarks successfully solved by all policies. Exact results of number of instances solved, average runtime, and median runtime in different policies appears in Appendix~\ref{section:appendixEvalutationStatisticsRank}, Fig~\ref{fig:comparePoliciesStats}.
	
	Analyzing the results in Fig.~\refsubref{fig:comparePolicies}{a} indicates that the \tool{}'s performance is not significantly affected by the choice of abstraction policy. Still, we see that the \emph{Single Class} policy solved more instances than any other policy but also took more time to solve them. Fig.~\refsubref{fig:comparePolicies}{b} shows that, when considering the accumulated runtime for a specific set of instances, the Single Class policy scores high and is second only to Majority Class Vote. Due to its relative success in both metrics, we conclude that it is the most successful of the considered policies.
	
	\subsubsection{Comparing \tool{} to vanilla Marabou.}
	\label{section:toolVanillaComparison}
	
\begin{figure}
			\begin{center}
				\includegraphics[width=0.65\textwidth]{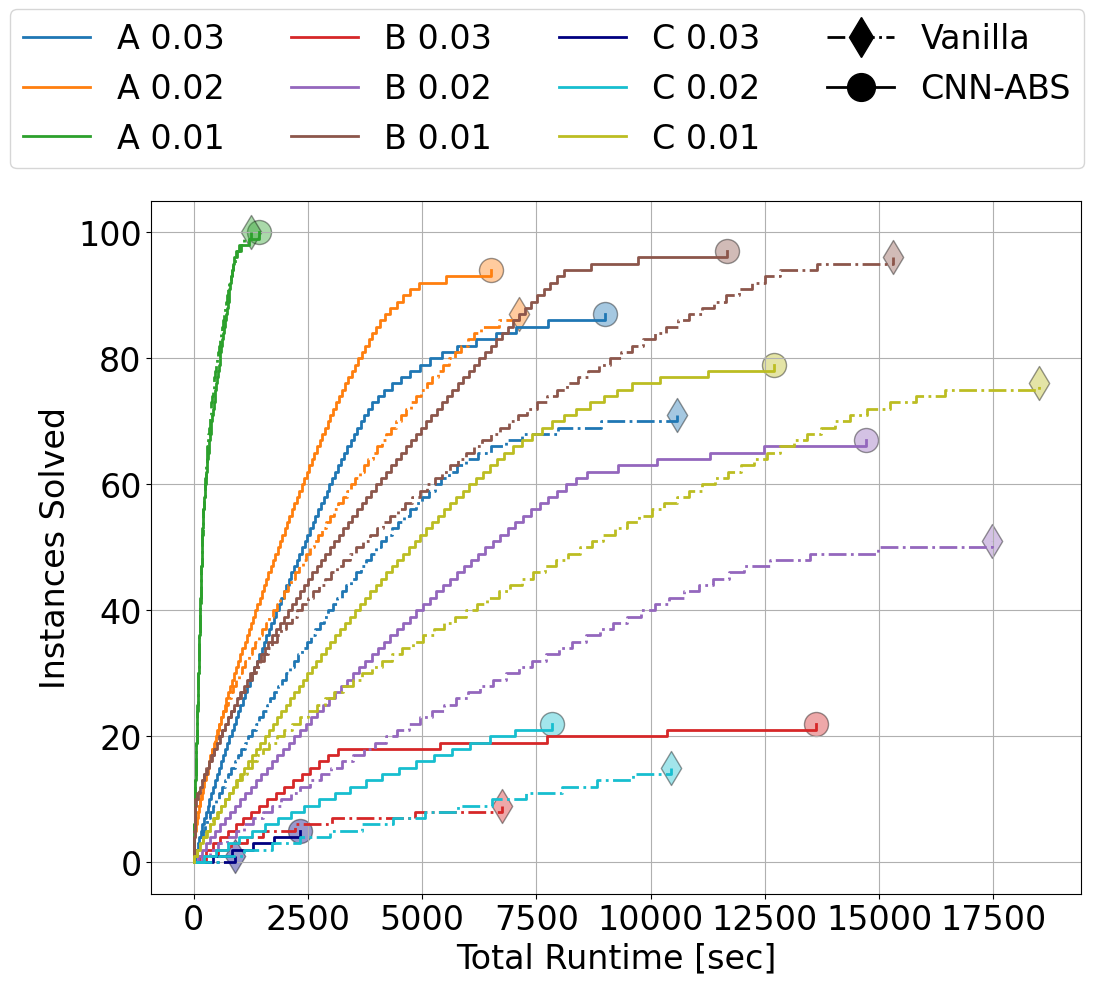}
			\end{center}
				\caption{Performance over the different networks ($A,B,C$) and values of $\varepsilon$ ($0.01,0.02,0.03$). Each query was ran in vanilla Marabou (dash-dotted line), and with \tool{} (solid line).}
				\label{fig:cacti}
	\end{figure}

	Next, we ran a comprehensive comparison between vanilla Marabou and our proof-of-concept implementation of \tool{}. For
	\tool{}, we abstract the deepest max-pooling layer using the \emph{Single Class} policy, which won in the first experiment. We used the same features as in the previous experiment but ran them on all three networks and with varying values of $\varepsilon$: $0.01,0.02$ and $0.03$. The result is nine combinations and a total of 900 experiments for each framework. The results are depicted in Fig.~\ref{fig:cacti}. Excluding the $(C,0.03),(B,0.03),(A,0.01)$ queries, in every category, the abstraction enhanced version solved more instances than vanilla and required a shorter total runtime. In the $(A,0.01)$ category, both frameworks performed similarly; and in $(C,0.03),(B,0.03)$, \tool{} solved more instances, but at the cost of additional runtime. Aggregating the results overall instances solved by both frameworks, \tool{}'s average runtime was 84.3\% that of vanilla Marabou's runtime, and its median runtime 75.4\% that of vanilla Marabou's. Additionally, \tool{} solved 1.13 times as many instances as vanilla Marabou. The exact numbers of instances solved, average runtimes, and median runtimes for each category appear in Appendix~\ref{section:appendixEvalutationStatisticsVanilla}, Fig~\ref{fig:compareVanillaCnnAbsStats}. This experiment clearly indicates the superior performance of \tool{} compared to the vanilla version.
	
\begin{figure}[t]
		\begin{minipage}[t]{0.45\textwidth}
			\vspace{0pt}
			\begin{center}
				\includegraphics[width=\textwidth]{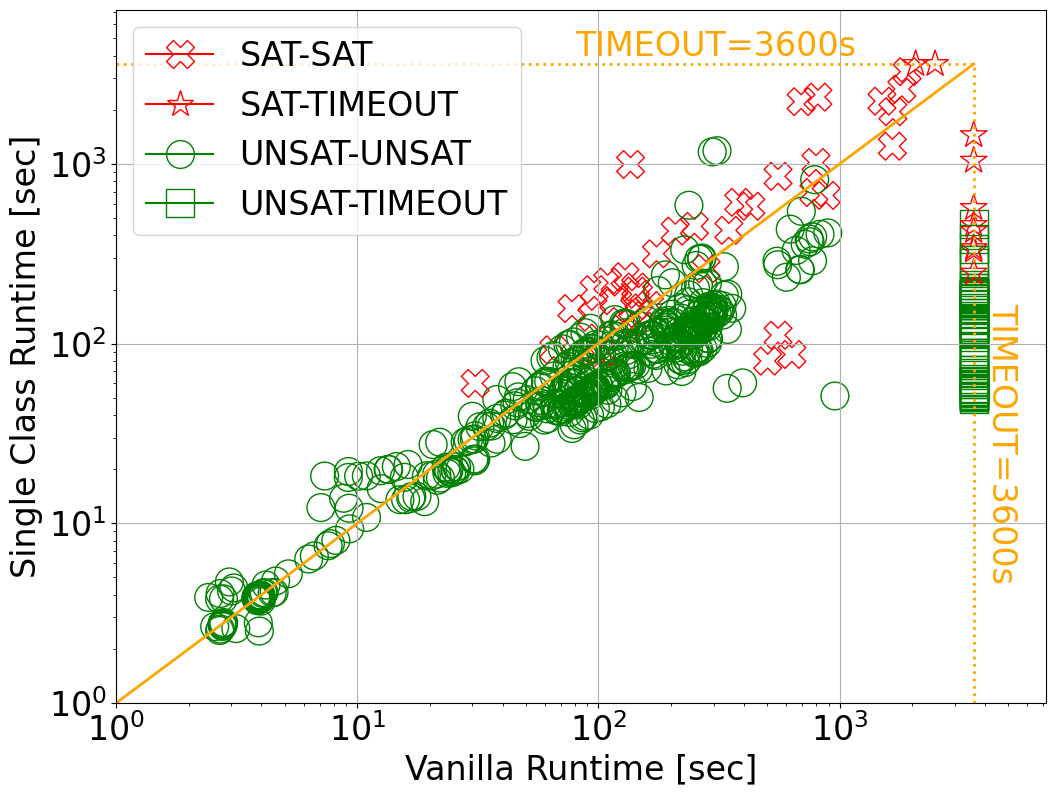}
				\caption{\tool{}'s runtime vs. vanilla Marabou's
					runtime, in log scale.}
				\label{fig:satvsunsat}
			\end{center}
		\end{minipage} \hspace{1mm}
		\begin{minipage}[t]{0.45\textwidth}
			\vspace{0pt}
			\begin{center}
				\includegraphics[width=\textwidth]{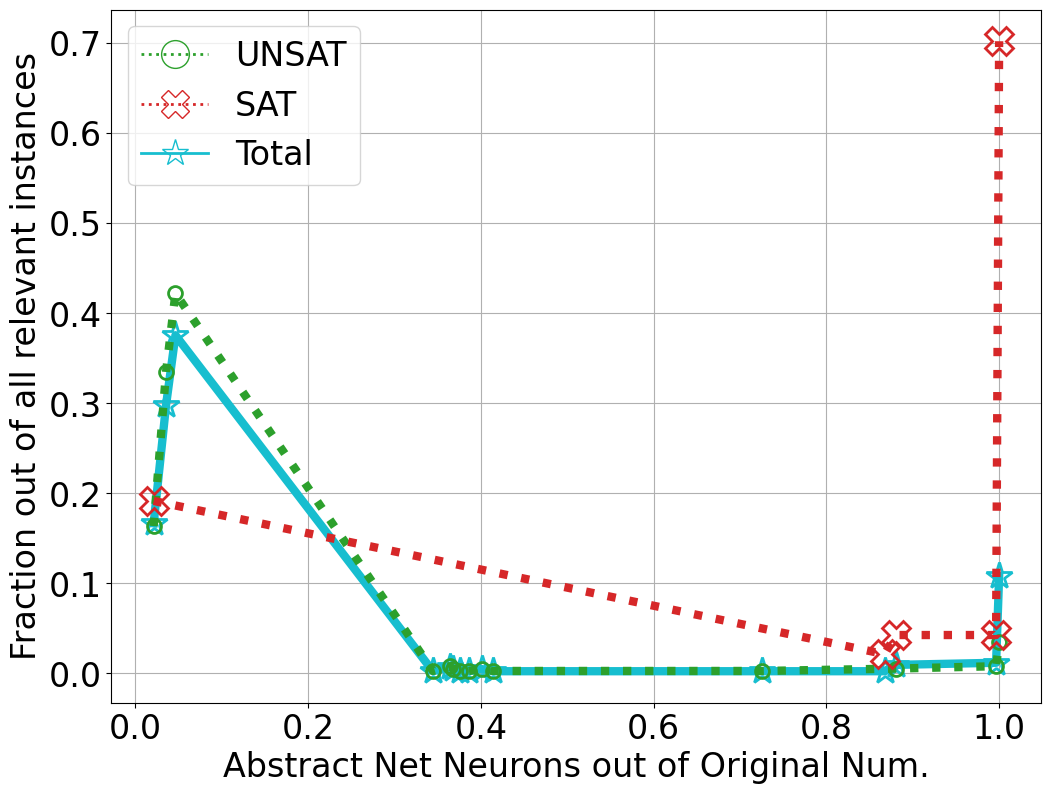}
				\caption{The sizes of the abstract
					networks when \tool{} terminates,
					compared to the original network.}
				\label{fig:hist}
			\end{center}
		\end{minipage}
	\end{figure}

	Fig.~\ref{fig:satvsunsat} depicts the runtime of \tool{} vs. vanilla Marabou, for every query solved by at least one of the verifiers. There are $526$ \unsat{} points (green) and $49$ \sat{} points (red). The results show that for \sat{} instances, the frameworks achieve similar performance; whereas for \unsat{} instances, \tool{} performs significantly better, solving 61 instances that the vanilla version timed out on. We thus conclude that the \tool{} is particularly effective on \unsat{} instances, presumably because \sat{} instances require multiple refinement steps.
	
	In Fig.~\ref{fig:hist}, we measure the number of refinement steps needed by \tool{} before arriving at an answer. Specifically, it depicts the size of the DNN in the final iteration of the abstraction/refinement algorithm as a fraction of the size of the original DNN. It is visible that the results mostly divide between \unsat{} queries, which terminate with small networks and few refinement steps, and \sat{} queries that often require the network to be refined back to the original DNN. The corollary is that slow, gradual refinement is ineffective; and that \tool{} performs better on \unsat{} queries, as these can often be solved on small, abstract networks.
	
	\section{Related Work}
	\label{section:Related Work}
	
	The topic of DNN verification has received significant
        attention in recent years, and many approaches have been
        proposed for addressing it. These include SMT-solving based
        approaches~\cite{Eh17, KaBaDiJuKo17, KaHuIbJuLaLiShThWuZe19,
          HuKwWaWu17, KuKaGoJuBaKo18}, reachability-based
        approaches~\cite{SiGePuVe19, GeMiDrTsChVe18, TjXiTe17, LoMa17,
          TrBaXiJo20}, abstraction-based approaches~\cite{PrAf20,
          ElGoKa20,AnPaDiCh19}, runtime verification and optimization
        approaches~\cite{LuScHe21,AvBlChHeKoPr19}, model-counting
        approaches~\cite{BaShShMeSa19} and many others; and these
        approaches have been applied in a variety of tasks, such as
        ensuring robustness~\cite{SiGePuVe19, KaBaDiJuKo17,
          CaKaBaDi17, GoKaPaBa18}, fairness~\cite{UrChWuZh20},
        modifying and simplifying DNNs~\cite{ReKa21, GoAdKeKa20,
          GoFeMaBaKa20, SoTh19, UsGoSuNoPa21}, augmenting deep reinforcement
        learning~\cite{AlAvHeLu20, ElKaKaSc21, KaBaKaSc19}, and
        beyond.  Generally, these approaches deal with verifying
        non-convolutional networks; but because convolution layers are
        a special case of weighted sum layers and the max functions
        are piecewise-linear, CNNs can usually be verified using these
        approaches.
	
	Some work has explicitly targeted CNN verification, proposing reachability-based approaches, either complete~\cite{KoLo18} or incomplete~\cite{BoWeChLiDa19, WuZh21, TrBaXiJo20}. In \cite{YaYaTrHoJoPr21}, an under-approximation reachability analysis method is presented. Some of these approaches use linear approximations and could benefit from our tighter approximation of the max function.  In contrast to these methods above, our technique leverages the special structure and connectivity of CNNs to reduce the size of their encoding and simplify their verification; and consequently, it can be integrated with many existing techniques.
	
	In a recent paper~\cite{XuLiZhDu21}, Xu et al.~take an approach similar to ours, and propose an SMT-based algorithm named Conv-Reluplex. Conv-Reluplex operates by splitting the network into two sub-networks, one of which is convolutional and the other fully connected; and it then verifies properties defined on the network's hidden layer that is on the border between the two sub-networks.  In contrast, our technique verifies properties given in the standard form~\cite{BaLiJo21}, i.e. properties given over the network's inputs. Exploring synergies between the two approaches is left for future work.
	
	The two novel components in our approach and framework are an abstraction-refinement based approach for CNN verification and a new bound tightening technique for max-pooling layers. Abstraction-refinement techniques have been successfully used to verify various kinds of systems~\cite{ClGrJhLuVe00, BoPaGi08}; and they have also been recently applied in the context of DNN verification~\cite{ElGoKa20, AsHaKrMo20, PrAf20}. Work so far has focused on merging neurons in order to produce a smaller, abstract network, whereas our approach focuses on removing edges and then pruning unneeded neurons entirely. Combining the edge-oriented and node-oriented abstraction approaches is left for future work. Bound tightening techniques for DNNs have been very extensively studied~\cite{Eh17, TjXiTe17, SiGePuVe19, DuJhSaTi18, WeZhChSoHsDaBoDh18, WaPeWhYaJa18, ZhWeChHsDa18, LyKoKoWoLiDa20, AmWuBaKa21, SaYaZhHsZh19, AnHuMaTjVi20, BoWeChLiDa19, WuZh21, ChNuRu17}, focusing mostly on the \relu{} function. Our work adds to this general line of research by proposing improved bounds for the max function.
	
	\section{Conclusion}
	\label{section:Conclusion}
	
	We presented a novel scheme for CNN verification, which uses abstraction-refinement techniques and bound tightening techniques tailored for max-pooling layers, which are common in CNNs. Our technique, \tool{}, is implemented in a proof-of-concept tool and can be used with various existing DNN verifiers as backends. We connected our tool to the Marabou verification engine and used the combined tool to demonstrate superior performance to those of vanilla Marabou. We regard this effort as a step towards more effective verification of real-world CNNs.
	
	Moving forward, we intend to pursue several directions. One direction is to attempt simultaneously abstracting neurons across multiple layers instead of a single layer at a time; another is to try and transform spurious counterexample into true ones by correcting the spurious values assigned to pruned neurons; yet another is to apply the approach also to non-convolutional networks.

	\subsubsection{Acknowledgements}
The project was partially supported by grants from the Binational
Science Foundation (2020250), the Israel Science Foundation (683/18),
and the Semiconductor Research Corporation.

	{
		\bibliographystyle{abbrv}
		\bibliography{cnnAbs}
	}
	
	\newpage
	\appendix
	\renewcommand{\thesection}{\Alph{section}}
	
	\section*{\huge Appendices}
	
	\section{Proof of used LP relaxation of the max function.}
	\label{section:appendixLP}
	
	Using the notation defined in Section~\ref{section:boundPropegation}
	\begin{gather*}
		b = max\{a_j\}_{j=0}^{k-1} \;\;,\;\; u_f = \uf \;\;,\;\; u_s = \us \\
		l_{max}=\lm \;\;,\;\; u_{min} = \um
	\end{gather*}
	
	when $s,f$ are the indices of the corresponding maximal elements, the bounds we use, as defined in Fig.\ref{equation:lpMaxNew}, are
	\begin{multline*}
		\bigwedge_{0\leq j \leq k-1}\left( b \geq a_j \right) \land \bigwedge_{\lambda \in \{l_{max},u_{min}\}}\left( b \leq \lambda + \sum_{i=0}^{k-1} \frac{\relu{}(u_i-\lambda)}{u_i-l_i}(a_i-l_i) \right) \land \\
		\left( b \leq u_f \frac{u_s - l_f}{u_f - l_f} + a_f \frac{u_f - u_s}{u_f-l_f} \right) 
	\end{multline*}
	when $u_s \geq l_{max}$, and otherwise $b = a_f$.
	
	\begin{figure}[H]
		\begin{center}
			\includegraphics[width=0.5\textwidth]{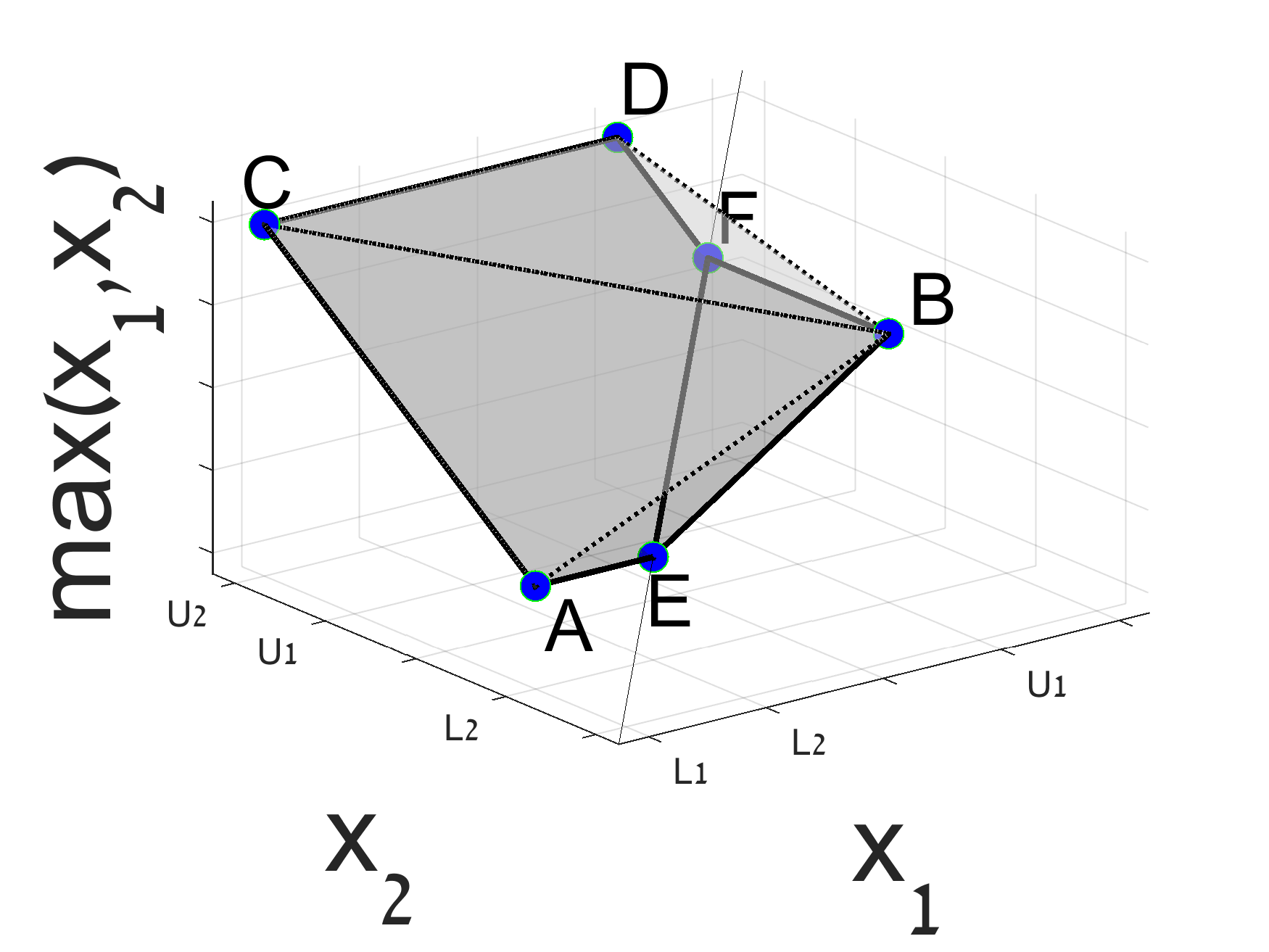}
		\end{center}
		\caption{Graph of the 2-dimensional max function and the faces of the convex hull. $AEFDC,BEF$ are the max function graph, and $ABC,BCD$ are faces of the convex-hull.}
		\label{fig:twodimmax}
	\end{figure}
	
	For convenience, we will refer to the bounds using $\lambda$ as $l_{max}$ and $u_{min}$ bounds, and the rightmost bound as the $u_f$ bound.
	
	The idea behind the bounds is phrasing the faces of the convex-hull of the $(k+1)$-dimensional max function graph. We take the 2-dimensional cases illustrated in Fig.~\ref{fig:twodimmax} as an example. The first conjunct visibly encodes the requirement of $max\{a_i\}$ to be larger than every $a_i$, but also encodes the graph faces $AEFDC, BEF$. The $l_{max}$ bound encodes the face $ABC$, and the $u_f$ bound encodes the $BCD$ face. In the 2D case, the $u_{min}$ bound coincide with the $BCD$ face, but this changes in higher dimensions.
	
	At the more general $(k+1)$-dimensional case, these bounds are the faces connecting the following vertices --- $k+1$-dimensional vectors where the $k+1$ coordinate is the max value of the first $k$ coordinates. One can see that each vertex set creates a $k$-dimensional affine space; therefore, the face, which is its orthogonal complement, is well defined as a $(k+1)$-dimensional hyperplane.
	
	\begin{itemize}
		\item The $u_f$ bound: encodes the face connecting the vertices
		\[
		(u_0,...,u_{k-1},u_f), \{(u_0,...,u_{i-1},l_i,u_{i+1},...,u_{k-1},max_{j\neq i}\{u_j\})\}_{i=0}^{k-1}
		\]
		The induced affine space is
		\[
		(u_0,...,u_{k-1},u_f) + span\left(\left\{(l_i-u_i)\cdot e_i + (max_{j\neq i}\{u_j\} - u_f)\cdot e_{k}\right\}_{i=0}^{k-1}\right)
		\]
		when $\{e_i\}_{i=0}^k$ is the $(k+1)$ standard basis.
		\item The $l_{max}$ bound: encodes the face connecting the vertices
		\[
		(l_0,...,l_{k-1},l_{max}), \{(l_0,...,l_{i-1},u_i,l_{i+1},...,l_{k-1},max\{l_{max}, u_i\}\}_{i=0}^{k-1}
		\] with the induced affine space 
		\[
		(l_0,...,l_{k-1},l_{max}) + span\left(\left\{(u_i-l_i)\cdot e_i + \relu(u_i-l_{max}) \cdot e_{k}\right\}_{i=0}^{k-1}\right)
		\]
		\item The $u_{min}$ bound: When $0 \leq \mu \leq k-1$ is an index different than $f$ such that $u_{\mu}=u_{min}$, the bound encodes the face connecting
		\begin{multline*}
			\{(l_0,...,l_{i-1},u_{i},l_{i+1},...,l_{k-1},u_{i})\}_{i = \mu,f},\\
			\{(l_0,...l_{i-1},u_i,l_{i+1},...,l_{\mu-1},u_{\mu},l_{\mu+1},...,l_{k-1},max(u_{i},u_{\mu}))\}_{i=0,i\neq \mu}^{k-1}
		\end{multline*}
		Notice that $u_f \geq l_{max}$ under our assumptions, therefore $u_{min}$ is well defined and bigger than all lower bounds. We can find $\mu\neq f$ because otherwise $u_s < l_{max}$, and this is again contradicting our assumption. The affine space induced by these vertices is
		\begin{multline*}
			(l_0,...,l_{\mu-1},u_{\mu},l_{\mu+1},...,l_{k-1},u_{\mu}) + \\
			span\left(\left\{(u_i-l_i)\cdot e_i + \relu(u_i-u_{\mu}) \cdot e_{k}\right\}_{i=0,i\neq \mu}^{k-1} \cup \{e_{\mu}\}\right)
		\end{multline*}
	\end{itemize}
	
	\subsection{Correctness of suggested bounds in Eq.~\ref{equation:lpMaxNew}}
	
	We additionally define the index of the maximal $a_j$ as $0 \leq M \leq k-1$ --- i.e $b = a_M$. Notice that if $u_s < l_{max}$ we can be sure that $b = a_f$, since $M$ must be equal to $f$ and $a_f$ will be above every upper bound of other variables. We will now assume that $l_{max} \leq u_s$, and need to prove the bound at Eq.~\ref{equation:lpMaxNew}.
	
	The lower bounds of $\bigwedge_{0\leq j \leq k-1}\left( b \geq a_j \right)$ are self evident because of the max function definition. First notice that $l_{max} \leq u_M$: Assume by contradiction that $l_{max} > u_M$, there would be index $m$ such that $a_m \geq l_m = l_{max} > u_M \geq a_M$ in contradiction to the maximality of $a_M$. This also means that $u_M$ participates in the set minimized by $u_{min}$, therefore $u_{min}$ well defined and $u_{min} \leq u_M$. By definition $l_{max} \geq l_M$, and $u_{min}\geq l_{max}$. Overall: $l_M \leq l_{max} \leq u_{min} \leq u_M$. We will prove the upper bounds. Let be $\lambda \in [l_{max},u_{min}]$ and consequently $\lambda \in [l_M,u_M]$. The following proves the upper bounds described at Eq.~\ref{equation:lpMaxNew}.
	
	\begin{flalign*}
		\lambda + \sum_{i=0}^{k-1} \frac{\relu{}(u_i-\lambda)}{u_i-l_i}(a_i-l_i) &\geq \\
		\text{\textit{(sum elements are non-negative)}} &\geq \lambda + \frac{\relu{}(u_M-\lambda)}{u_M-l_M}(a_M-l_M)\\
		\text{\textit{($u_M \geq \lambda$)}} &\geq \lambda + \frac{u_M-\lambda}{u_M-l_M}(a_M-l_M)\\
		&= \frac{\lambda(u_M - l_M) + (u_M-\lambda)(a_M-l_M)}{u_M-l_M}\\
		&= \frac{u_M(\lambda - l_M) + a_M (u_M - \lambda)}{u_M-l_M}\\
		\textit{($u_M \geq a_M$ and $\lambda \geq l_M$)} &\geq  \frac{a_M(\lambda - l_M) + a_M (u_M - \lambda)}{u_M-l_M} = a_M = b \;\;\square\\
	\end{flalign*}
	Now, prove the remaining part of the upper bounds.
	\begin{flalign*}
		u_f \frac{u_s - l_f}{u_f - l_f} + a_f \frac{u_f - u_s}{u_f-l_f} &=  \frac{u_f u_s -u_f l_f + a_f u_f - a_f u_s}{u_f-l_f} = \\
		&= u_s \frac{u_f - a_f}{u_f - l_f} + u_f \frac{a_f - l_f}{u_f - l_f}\\
		\textit{($u_f \geq u_s$)} &\geq u_s \frac{u_f - a_f}{u_f - l_f} + u_s \frac{a_f - l_f}{u_f - l_f} = u_s \\
	\end{flalign*}
	and
	\begin{flalign*}
		u_f \frac{u_s - l_f}{u_f - l_f} + a_f \frac{u_f - u_s}{u_f-l_f} &=  \frac{u_f u_s -u_f l_f + a_f u_f - a_f u_s}{u_f-l_f} = \\
		&= \frac{u_f u_s -u_f l_f + a_f u_f - a_f u_s + a_f l_f - a_f l_f}{u_f-l_f}\\
		&= a_f + (u_f-a_f)\frac{u_s-l_f}{u_f-l_f}\\
		\textit{($u_s \geq l_{max} \geq l_f$)} &\geq a_f\\
	\end{flalign*}
	Altogether we get 
	\[
	u_f \frac{u_s - l_f}{u_f - l_f} + a_f \frac{u_f - u_s}{u_f-l_f} \geq a_f,u_s
	\]
	and considering that $\forall j \neq f \; u_s \geq u_j \geq a_j$, we get that the bound is bigger than all $a_j$, among them $a_M = b$. $\square$
	
	\subsection{Definition Of State-Of-The-Art}
	
	For $\gamma_0 = \gammazero$ and $\gamma = min(max(\gamma_0, l_{max}), u_{min})$ , we refer to the state-of-the-art as 
	\begin{multline}
		\bigwedge_{0\leq j \leq k-1}\left( b \geq a_j \right) \land \left( b \leq \gamma + \sum_{i=0}^{k-1} \frac{\relu{}(u_i-\gamma)}{u_i-l_i}(a_i-l_i) \right) \land \\ 
		\left( b \leq u_f \right)  \land \left( b \leq l_{max} + \sum_{i=0}^{k-1}(a_i-l_i)\right)
		\label{equation:stateoftheartLP}
	\end{multline}
	when $l_{max} \leq u_s$, and $b=a_f$ otherwise. This expression is a combination of the following bounds:
	
	\begin{itemize}
		\item The bound presented at Planet~\cite{Eh17}:
		\begin{equation}
			\bigwedge_{0\leq j \leq k-1}\left( b \geq a_j \right) \land \left( b \leq l_{max} + \sum_{i=0}^{k-1}(a_i-l_i)\right)
			\label{equation:plantLP}
		\end{equation}
		\item The bound presented at Deeppoly~\cite{SiGePuVe19}:
		If $u_s < l_{max}$ then $b = a_f$. Else,
		\begin{equation}
			\left(b \geq a_m \right) \land \left(b \leq u_f \right)
			\label{equation:deeppolyLP}
		\end{equation}
		when $0\geq m \geq k-1$ such that $l_m = l_{max}$.
		\item The bound presented at CNN-CERT~\cite{BoWeChLiDa19}:
		Define
		\[
		\gamma_0 = \gammazero \;\;,\;\; \gamma = min(max(\gamma_0, l_{max}), u_{min}) \;\;,\;\;  G = \sum_{i=0}^{k-1}\frac{u_i-\gamma}{u_i-l_i}
		\]
		\[
		\eta =  \left\{\begin{array}{lc}
			min_{i=0}^{k-1}\{l_i\} & \text{if } G < 1\\
			max_{i=0}^{k-1}\{u_i\} & \text{if } G > 1\\
			\gamma                 & \text{if } G = 1
		\end{array}\right.\;\;,\;\; \gamma_0 = \gammazero
		\]
		The bound is
		\begin{equation}
			\left( b \geq \eta + \sum_{i=0}^{k-1} \frac{\relu{}(u_i-\gamma)}{u_i-l_i}(a_i-\eta) \right) \land \left( b \leq \gamma + \sum_{i=0}^{k-1} \frac{\relu{}(u_i-\gamma)}{u_i-l_i}(a_i-l_i) \right)
			\label{equation:cnncertLP}
		\end{equation}
	\end{itemize}
	
	The state-of-the-art lower bounds in Eq.~\ref{equation:stateoftheartLP} are the ones presented in Planet's Eq.~\ref{equation:plantLP}: $\bigwedge_{0\leq j \leq k-1}\left( b \geq a_j \right)$. They are the optimal tightest lower bounds, since any lower bound $L(a_1,...,a_k)$ by definition will maintain $L(a_1,...,a_k) \leq a_M \leq b$. The state-of-the-art bounds include the intersection of all $b \geq a_i$, when specifically $b \geq a_M$ is included and this is the tightest possible linear bound.
	The state-of-the-art upper bound is the intersection of the three bounds. Being the intersection of all, it is trivially tighter than any of them separately. $\square$
	
	\subsection{Improvement Over State-Of-The-Art}
	
	Our suggested bounds, as described in Section~\ref{section:boundPropegation}, Eq.~\ref{equation:lpMaxNew}, are:
	\begin{multline}
		\bigwedge_{0\leq j \leq k-1}\left( b \geq a_j \right) \land \bigwedge_{\lambda \in \{l_{max},u_{min}\}}\left( b \leq \lambda + \sum_{i=0}^{k-1} \frac{\relu{}(u_i-\lambda)}{u_i-l_i}(a_i-l_i) \right) \land \\
		\left( b \leq u_f \frac{u_s - l_f}{u_f - l_f} + a_f \frac{u_f - u_s}{u_f-l_f} \right) 
	\end{multline}
	when $l_{max} \leq u_s$, and $b=a_f$ otherwise.
	
	When $l_{max} > u_s$, choosing $b=a_f$ is the optimal bound, and this is the same as the state-of-the-art. Now, assume $l_{max} \leq u_s$. The lower bound is identical to this of the state-of-the-art. We will prove that our upper bounds are tighter than the state-of-the-art. It should be said that these bounds are the tightest we encountered, while we allow ourselves to use more than a single lower and upper linear inequality. The bounds mentioned at Eq.~\ref{equation:deeppolyLP} Eq.~\ref{equation:cnncertLP} were built with that limitation, and the jury is still out on the tightest bound maintaining that demand.
	
	Take our bounds of
	\begin{equation}
		\bigwedge_{\lambda \in \{l_{max},u_{min}\}}\left( b \leq \lambda + \sum_{i=0}^{k-1} \frac{\relu{}(u_i-\lambda)}{u_i-l_i}(a_i-l_i) \right)
		\label{equation:upperGammaBounds}
	\end{equation}
	For all $b,\{a_i\}_{i=0}^{k-1}$
	\begin{flalign*}
		l_{max} + \sum_{i=0}^{k-1} \frac{\relu{}(u_i-l_{max})}{u_i-l_i}(a_i-l_i) &\leq \\
		\text{\textit{($l_{max}\geq l_i$)}} &\leq l_{max} + \sum_{i=0}^{k-1} \frac{\relu{}(u_i-l_i)}{u_i-l_i}(a_i-l_i) \\
		&= l_{max} + \sum_{i=0}^{k-1} (a_i-l_i) 
	\end{flalign*}
	and therefore our bound is tighter than the $l_{max} + \sum_{i=0}^{k-1} (a_i-l_i)$ term. Regarding the $b \leq \gamma + \sum_{i=0}^{k-1} \frac{\relu{}(u_i-\gamma)}{u_i-l_i}(a_i-l_i)$ term, notice that $l_{max} \leq \gamma \leq u_{min}$ and that the structure of the bounds is the same excluding the changing parameter. 
	
	We will prove a move general statement: taking some $\lambda \in [l_{max},u_{min}]$, the bounds in Eq.~\ref{equation:upperGammaBounds} are tighter than $b \leq \lambda + \sum_{i=0}^{k-1} \frac{\relu{}(u_i-\lambda)}{u_i-l_i}(a_i-l_i)$. Demand 
	\begin{flalign*}
		l_{max} + \sum_{i=0}^{k-1} \frac{\relu{}(u_i-l_{max})}{u_i-l_i}(a_i-l_i)    &\leq \lambda + \sum_{i=0}^{k-1}  \frac{\relu{}(u_i-\lambda)}{u_i-l_i}(a_i-l_i) \\
		&\iff \\
		\sum_{u_i\geq l_{max}} \frac{a_i-l_i}{u_i-l_i}(\lambda-l_{max}) &\leq \lambda - l_{max} \\
		&\iff (\text{\textit{or $\lambda = l_{max}$ and the bounds are equivalent}})\\
		\sum_{u_i\geq l_{max}} \frac{a_i-l_i}{u_i-l_i} &\leq 1 \\
	\end{flalign*}
	Similarly for the $u_{min}$ bound
	\begin{flalign*}
		u_{min} + \sum_{i=0}^{k-1} \frac{\relu{}(u_i-u_{min})}{u_i-l_i}(a_i-l_i)    &\leq \lambda + \sum_{i=0}^{k-1}  \frac{\relu{}(u_i-\lambda)}{u_i-l_i}(a_i-l_i) \\
		&\iff \\
		\sum_{u_i\geq l_{max}} \frac{a_i-l_i}{u_i-l_i}(\lambda-u_{min}) &\leq \lambda - u_{min} \\
		&\iff (\text{\textit{or $\lambda = u_{min}$ and the bounds are equivalent}})\\
		\sum_{u_i\geq l_{max}} \frac{a_i-l_i}{u_i-l_i} &\geq 1 \\
	\end{flalign*}
	Overall, for any $\sum_{u_i\geq l_{max}} \frac{a_i-l_i}{u_i-l_i}$ value one of the $l_{max},u_{min}$ bounds in Eq.~\ref{equation:upperGammaBounds} is little or equal to the $\lambda$ bound, and their intersection --- that we use --- is tighter than the $\lambda$ bound. Following this proof, our bound in Eq.~\ref{equation:upperGammaBounds} is tighter than the $\gamma$ bound in Eq.~\ref{equation:stateoftheartLP} since $\gamma \in [l_{max},u_{min}]$.
	
	Finally, we can see that 
	\begin{flalign*}
		u_f \frac{u_s - l_f}{u_f - l_f} + a_f \frac{u_f - u_s}{u_f-l_f} &= \\ 
		&= a_f + (u_f-a_f)\frac{u_s-l_f}{u_f-l_f} \\
		\textit{($\frac{u_s-l_f}{u_f-l_f} \leq 1$)} &\leq a_f + (u_f-a_f) = u_f\\
	\end{flalign*}
	This concludes the proof, and our suggested bound at Eq.~\ref{equation:lpMaxNew} is tighter than the state-of-the-art in Eq.~\ref{equation:stateoftheartLP}. $\square$
	
	\subsection{State-Of-The-Art LP Relaxation of Fig~\ref{fig:LPExample}}
	
	Considering the incoming bound to the max neurons, $c_0 \in [-2,2], c_1 \in [-3,3], c_2 \in [-4,4]$, we get:
	\begin{flalign*}
		l^{m_0}_{max} = -2 \;\;,\;\; u^{m_0}_{min} = 2 \;\;,\;\; \gamma_{0}^{m_0} = 0 \;\;,\;\; \gamma^{m_0} = 0\\
		l^{m_1}_{max} = -3 \;\;,\;\; u^{m_1}_{min} = 3 \;\;,\;\; \gamma_{0}^{m_1} = 0 \;\;,\;\; \gamma^{m_1} = 0\\
	\end{flalign*}
	
	The LP formulation of the state-of-the-art bounds in Eq.~\ref{equation:stateoftheartLP} is displayed in Fig.~\ref{fig:stateoftheartLP}. The maximization yields an upper bound of $y\leq 7$.
	
	\begin{figure}[h]
		\begin{minipage}[t]{0.5\textwidth}
			\vspace{0pt}
			\begin{center}
				\textbf{LP Query:}\\
				\textbf{Maximize $y$\ s.t.}
				
				$\left.\begin{array}{l}
					-1\leq x_0,x_1 \leq 1 \\
					-2\leq x_2,x_3 \leq 2 \\
				\end{array}\right\} \begin{array}{l} \text{Input} \\ \text{constraints}\end{array}$\\
				$\left.\begin{array}{l}
					c_0=x_0 - x_1 \\
					c_1=x_1 - x_2 \\
					c_2=x_2 - x_3 \\	
				\end{array}\right\} \begin{array}{l} \text{2nd layer} \\ \text{connections}\end{array}$\\
			\end{center}
		\end{minipage}
		\begin{minipage}[t]{0.5\textwidth}
			\vspace{0pt}
			\begin{center}
				$\left.\begin{array}{l}
					m_0 >= c_0,c_1 \\
					m_0 - \frac{1}{2}c_0 - \frac{1}{2}c_1 <= 2.5  \\
					m_0 - c_0 - c_1 <= 3 \\
					m_0 <= 3 \\
					m_1 >= c_1,c_2 \\
					m_1 - \frac{1}{2}c_1 - \frac{1}{2}c_2 <= 3.5  \\
					m_1 - c_1 - c_2 <= 4 \\
					m_1 <= 4 \\
				\end{array}\right\} \begin{array}{l} \text{Max} \\ \text{functions}\end{array}$\\
				$\left.\begin{array}{l}
					y=m_0 + m_1 \\
				\end{array}\right\} \begin{array}{l} \text{4th layer} \\ \text{connections}\end{array}$\\
				\textbf{Result: y = 7}
			\end{center}
		\end{minipage}
		\caption{State-Of-The-Art LP Relaxation.}
		\label{fig:stateoftheartLP}
	\end{figure}
	
	\section{Detailed Structure of Networks Used During Evaluation}
	\label{section:appendixNetworkStructure}
	
	Here, in Fig.~\ref{fig:networkStructure} we present the structure of the networks used in Section~\ref{section:Evaluation}.
	
	\begin{figure}[H]
		\begin{minipage}[t]{0.3\textwidth}
			\begin{center}
				$\begin{array}{l c}
					\textbf{Network A} & \\
					\hline
					\text{Accuracy: }     & 93.7\%\\
					\text{Num. Neurons: } & 2719 \\
					\hline
					\multicolumn{2}{c}{\textbf{Layers}} \\
					\textbf{Type}   & \textbf{Dim.} \\
					\hline \\
					Input                 &\mat{28 & 28 & 1} \\
					Convolution           &\mat{26 & 26 & 1} \\
					\relu                 &\mat{26 & 26 & 1} \\
					Max-Pooling           &\mat{13 & 13 & 1} \\
					Convolution           &\mat{12 & 12 & 1} \\
					\relu                 &\mat{12 & 12 & 1} \\
					Max-Pooling           &\mat{6 & 6 & 1} \\
					WS                    &\mat{40} \\
					\relu                 &\mat{40} \\
					Output                &\mat{10} \\
					&\\
					&\\
					&\\
				\end{array}$
				\label{fig:networkStructure-a}
			\end{center}
		\end{minipage}\hspace{0.025\textwidth}\textcolor{lightgray}{\vline}\hspace{0.0125\textwidth}
		\begin{minipage}[t]{0.3\textwidth}
			\begin{center}
				$\begin{array}{l c}
					\textbf{Network B} & \\
					\hline
					\text{Accuracy: }     & 96.2\%\\
					\text{Num. Neurons: } & 4564 \\
					\hline
					\multicolumn{2}{c}{\textbf{Layers}} \\
					\textbf{Type}   & \textbf{Dim.} \\
					\hline \\
					Input                 &\mat{28 & 28 & 1} \\
					Convolution           &\mat{26 & 26 & 2} \\
					\relu                 &\mat{26 & 26 & 2} \\
					Max-Pooling           &\mat{13 & 13 & 2} \\
					Convolution           &\mat{12 & 12 & 2} \\
					\relu                 &\mat{12 & 12 & 2} \\
					Max-Pooling           &\mat{6 & 6 & 2} \\
					WS                    &\mat{40} \\
					\relu                 &\mat{40} \\
					Output                &\mat{10} \\
				\end{array}$
				\vspace{1cm}
				\label{fig:networkStructure-b}
			\end{center}
		\end{minipage}\hspace{0.025\textwidth}\textcolor{lightgray}{\vline}\hspace{0.0125\textwidth}
		\begin{minipage}[t]{0.3\textwidth}
			\begin{center}
				$\begin{array}{l c}
					\textbf{Network C} & \\
					\hline
					\text{Accuracy: }     & 86.6\%\\
					\text{Num. Neurons: } & 4636 \\
					\hline
					\multicolumn{2}{c}{\textbf{Layers}} \\
					\textbf{Type}   & \textbf{Dim.} \\
					\hline \\
					Input                 &\mat{28 & 28 & 1} \\
					Convolution           &\mat{26 & 26 & 2} \\
					\relu                 &\mat{26 & 26 & 2} \\
					Max-Pooling           &\mat{13 & 13 & 2} \\
					Convolution           &\mat{12 & 12 & 2} \\
					\relu                 &\mat{12 & 12 & 2} \\
					Max-Pooling           &\mat{6 & 6 & 2} \\
					Convolution           &\mat{4 & 4 & 2} \\
					\relu                 &\mat{4 & 4 & 2} \\
					Max-Pooling           &\mat{2 & 2 & 2} \\
					WS                    &\mat{40} \\
					\relu                 &\mat{40} \\
					Output                &\mat{10} \\
				\end{array}$
				\label{fig:networkStructure-c}
			\end{center}
		\end{minipage}
		\caption{Networks used in evaluation. For each network accuracy, number of neurons, and layers.}
		\label{fig:networkStructure}
	\end{figure}
	
	\section{Detailed Statistics Regarding Performed Experiments}
	\label{section:appendixEvalutationStatistics}
	
	\subsection{Ranking Policy Comparison}
	\label{section:appendixEvalutationStatisticsRank}
	
	The table at Fig.~\ref{fig:comparePoliciesStats} compares the results of different policies for the evaluation done in Section~\ref{section:rankPolicies}, and specifically Fig.~\ref{fig:comparePolicies}. We can see that the median runtime is fairly similar in all policies. Looking at the Single Class policy, we can see it scored the top regarding the number of solved instances. In combination with \refsubref{fig:comparePolicies}{b}, it appears that the added number of instances, which are probably relatively hard since they were not solved by other policies, took relativity more runtime and increased its average runtime over other policies.
	\begin{figure}
		\begin{center}
			\begin{tabular}{||l||c|c|c||}
				\hline\hline
				Policy   & Instances solved (out of 100)& average runtime& median runtime\\
				\hline\hline
				Centered            & 75 & $225s$ & $60s$ \\
				\hline
				All Samples         & 75 & $222s$ & $58s$ \\
				\hline
				Sample Rank         & 75 & $271s$ & $61s$ \\
				\hline
				Single Class        & 77 & $289s$ & $59s$ \\
				\hline
				Majority Class Vote & 75 & $228s$ & $57s$ \\
				\hline
				Random              & 73 & $184s$ & $56s$ \\
				\hline
			\end{tabular}
		\end{center}
		\caption{Performance of ranking policy specified in Section~\ref{section:rankPolicies} for results shown at Fig.~\ref{fig:comparePolicies}.}
		\label{fig:comparePoliciesStats}
	\end{figure}
	
	\subsection{Comparing \tool{} to Vanilla Marabou}
	\label{section:appendixEvalutationStatisticsVanilla}
	
	\subsubsection{Runtime Comparison}
	
	The table at Fig.~\ref{fig:compareVanillaCnnAbsStats} compares the results of \tool{} vs. vanilla Marabou done in Section~\ref{section:toolVanillaComparison} and also presented at Fig.~\ref{fig:cacti}. The results show \tool{} improves the number of solved instances and average and median runtimes for all queries excluding $(A,0.01)$. This query seems to saturate, quickly solving all of the instances. Its median runtime is equal for \tool{} and vanilla Marabou, and the average runtime is increased in only $2$ second in \tool{}. For this reason, $(A,0.01)$ is not significant in the overall analysis, and combined with the positive results for the other queries, the performance improvement in \tool{} is significant. 
	
	We performed an additional analysis of all instances, not separating the different configurations. Considering samples that were successfully solved by both \tool{} and vanilla Marabou, on average \tool{}'s runtime was 84.3\% of vanilla Marabou runtime, while the median result was 75.4\%. This result is added to \tool{} solving 1.13 times the instances that vanilla Marabou did: 573 out of 900 for \tool{}, and 506 for vanilla Marabou.
	
	\begin{figure}
		\begin{center}
			\begin{tabular}{||l||c|c||c|c||c|c||}
				\hline\hline
				Query    & VM solved & \textsc{Cnn-A}. solved& VM avg.& \textsc{Cnn-A} avg.& VM med.& \textsc{Cnn-A} med.\\
				\hline\hline
				(A,0.01) & 100& 100& $12s$ & $14s$ & $4s$  & $4s$  \\
				\hline
				(A,0.02) & 87 & 94 & $82s$ & $69s$ & $81s$ & $49s$ \\
				\hline
				(A,0.03) & 71 & 87 & $149s$& $104s$& $94s$ & $56s$ \\
				\hline
				(B,0.01) & 96 & 97 & $159s$& $120s$& $139s$& $95s$ \\
				\hline
				(B,0.02) & 51 & 67 & $343s$& $220s$& $262s$& $137s$\\
				\hline
				(B,0.03) & 9  & 22 & $749s$& $619s$& $357s$& $185s$\\
				\hline
				(C,0.01) & 76 & 79 & $243s$& $161s$& $233s$& $110s$\\
				\hline
				(C,0.02) & 15 & 22 & $696s$& $357s$& $694s$& $334s$\\
				\hline
				(C,0.03) & 1  & 5  & $893s$& $465s$& $893s$& $458s$\\
				\hline
				Total    & 506  & 573  & & & & \\
				\hline
			\end{tabular}
		\end{center}
		\caption{Performance of \tool{} compared to vanilla Marabou on queries specified in Section~\ref{section:toolVanillaComparison} for results shown at Fig.~\ref{fig:cacti}. Number of solved instances (out of 100), average runtime per sample, and median runtime per sample.}
		\label{fig:compareVanillaCnnAbsStats}
	\end{figure}
	
	\subsubsection{Required Abstraction Steps}
	
	The table in Fig.~\ref{fig:AbstractionSteps} shows the abstraction status of the successful verification queries for the experiment shown in Fig.~\ref{fig:cacti}. The options are that a query is considered infeasible (\unsat{}) during the initial LP bound tightening, or that it was successfully solved with all neruons in the abstracted layer being abstracted, with some neurons being abstracted, or with none being abstracted when the original network is used following a full refinement.

	Most of the queries resulting in \unsat{} are solved during the initial LP bound tightening --- by proving the query is infeasible --- or by the first abstraction attempt in which all chosen layer's neurons are abstracted. For the \sat{} queries most successes were achieved after fully refining the query and verifying the original network.
	We can see that in some of the queries for both \sat{} and \unsat{} results \tool{} solved on an abstract network after some refinement steps. This is important since these are the queries in which the abstraction policies differ. We can see that are few queries solved in said status, but this is still a small portion of the entirety of solved queries.
	
	\begin{figure}
		\begin{center}
			\begin{tabular}{||l||c|c|c|c||c||}
				\hline\hline
				&  LP \unsat{} & All Neurons Abs. & Abs. With Some Refinement & Full Network & Total\\
				\hline\hline
				\unsat{}     & 147 & 349 & 17  & 13 & 526 \\
				\hline
				\sat{}       & 0   & 9   & 5   & 33 & 47  \\
				\hline
				Total        & 147 & 358 & 22  & 46 & 573 \\
				\hline
			\end{tabular}
		\end{center}
		\caption{Amount of abstracted neurons inside the abstracted layer in successful \tool{} verification queries for results shown at Fig.~\ref{fig:cacti}}
		\label{fig:AbstractionSteps}
	\end{figure}
	
\end{document}